\pdfoutput=1

\documentclass{article}

\usepackage{microtype}
\usepackage{graphicx}
\usepackage{subcaption}
\usepackage{booktabs} 

\usepackage{hyperref}
\usepackage{placeins}
\usepackage{enumitem}

\newcommand{\ourmethod}{SCALAR}

\usepackage{float}

\usepackage[preprint]{icml2026}

\usepackage{amsmath}
\usepackage{amssymb}
\usepackage{mathtools}
\usepackage{amsthm}
\usepackage{enumitem}
\usepackage{wrapfig}
\usepackage[most]{tcolorbox}
\usepackage{listings}
\usepackage[capitalize,noabbrev]{cleveref}

\theoremstyle{plain}

\theoremstyle{definition}

\theoremstyle{remark}

\usepackage[disable,textsize=tiny]{todonotes}

\newtcolorbox{skillbox}[2][]{%
  colback=gray!2, colframe=black!60, boxrule=0.6pt, arc=2pt,
  left=6pt,right=6pt,top=5pt,bottom=5pt,
  title=\textbf{#2}, fonttitle=\bfseries\footnotesize,
  fontupper=\footnotesize,
  #1
}

\icmltitlerunning{Learning and Composing Skills through LLM Guided Symbolic Planning and Deep RL Grounding}

\begin{document}

\twocolumn[
  \icmltitle{SCALAR: Learning and Composing Skills \\ through LLM Guided Symbolic Planning and Deep RL Grounding}



  \icmlsetsymbol{equal}{*}

  \begin{icmlauthorlist}
    \icmlauthor{Renos Zabounidis}{cmu}
    \icmlauthor{Yue Wu}{cmu}
    \icmlauthor{Simon Stepputtis}{vt}
    \icmlauthor{Woojun Kim}{cmu}
    \icmlauthor{Yuanzhi Li}{cmu}
    \icmlauthor{Tom Mitchell}{cmu}
    \icmlauthor{Katia Sycara}{cmu}
  \end{icmlauthorlist}

  \icmlaffiliation{cmu}{School of Computer Science, Carnegie Mellon University}
  \icmlaffiliation{vt}{Department of Mechanical Engineering, Virginia Tech}
  \icmlcorrespondingauthor{Renos Zabounidis}{renosz@andrew.cmu.edu}

  \icmlkeywords{Machine Learning, ICML}

  \vskip 0.3in
]



\printAffiliationsAndNotice{}  

\begin{abstract}
LLM-based agents excel when given high-level action APIs but struggle to ground language into low-level control. Prior work has LLMs generate skills or reward functions for RL, but these one-shot approaches lack feedback to correct specification errors. We introduce SCALAR, a bidirectional framework coupling LLM planning with RL through a learned skill library. The LLM proposes skills with preconditions and effects; RL trains policies for each skill and feeds back execution results to iteratively refine specifications, improving robustness to initial errors. Pivotal Trajectory Analysis corrects LLM priors by analyzing RL trajectories; Frontier Checkpointing optionally saves environment states at skill boundaries to improve sample efficiency. On Craftax, SCALAR achieves 88.2\% diamond collection, a 1.9$\times$ improvement over the best baseline, and reaches the Gnomish Mines 9.1\% of the time where prior methods fail entirely.
\end{abstract}

\section{Introduction}

An open challenge in creating LLM-based agents is grounding their actions in the world they interact with. Despite remarkable success in complex domains~
\citep{chowdhery2022palm,openai2023gpt4,touvron2023llama,o1,r1}, even sophisticated agents interact with environments by writing code rather than direct control. Voyager~
\citep{wang2023voyager} learns to craft items and explore Minecraft by generating code for the Mineflayer~
\citep{PrismarineJS_mineflayer_2026} API. The Factorio Learning Environment has LLMs write factory automation scripts, with low-level execution handled by the underlying system~
\citep{hopkins2025factoriolearningenvironment}. These systems demonstrate sophisticated reasoning, but abstract away the low-level control problem entirely.
\begin{figure}[H]
  \centering
\includegraphics[width=\columnwidth]{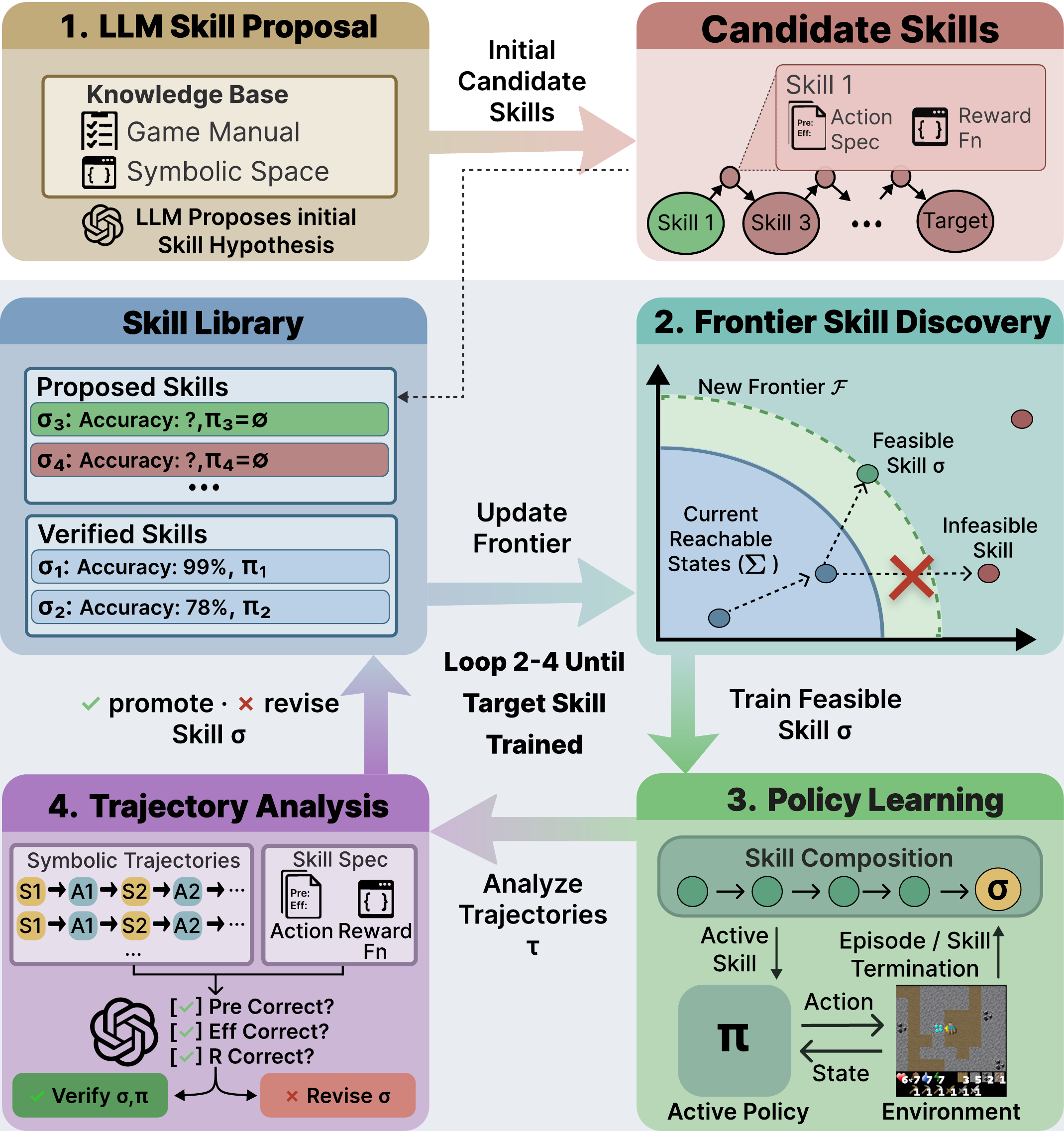}
  \caption{Overview of SCALAR's LLM-RL loop. The LLM reads the game manual and proposes candidate skills as symbolic operators with preconditions and effects (\textbf{1}). Skills whose preconditions are reachable from the current frontier are marked feasible and added to the skill graph; infeasible skills are deferred (\textbf{2}). Given a target skill, a planner sequences existing skills to reach a state satisfying its preconditions; an RL agent trains a policy for the target skill by executing these prerequisite skills and then learning from its own exploration (\textbf{3}). After training, trajectory analysis examines successful executions to verify or correct the LLM's initial specification of preconditions and resource requirements (\textbf{4}). Revised specifications update the skill graph and inform subsequent LLM proposals. This loop repeats until the target skill achieves sufficient success rate.}
  \label{fig:scalar_diamond}
\end{figure}

\noindent This abstraction enables impressive high-level behavior, yet prevents these agents from acquiring motor skills that require direct interaction~\citep{dai2026languagemovementprimitivesgrounding}.

\citet{li2024larm} quantify this limitation: when LLMs can write code for high-level APIs, they craft complex items over half the time, but when restricted to low-level motor control, the same approach achieves only 70\% success on basic items and fails entirely at complex tasks. The issue is not task difficulty but control granularity: code generation lets LLMs leverage common-sense planning, while low-level motor control requires skills beyond their capabilities. Conversely, RL can learn motor control through trial and error, but faces intractable exploration when rewards are sparse across long action sequences. The challenge is to define tasks at the right granularity: structured enough that LLMs can specify them from prior knowledge, yet short-horizon enough that RL can learn them with reasonable exploration. We call such tasks \emph{skills}.

Skills are short-horizon behaviors with clear success criteria that provide RL with focused objectives and intermediate feedback, while giving LLMs a vocabulary for hierarchical reasoning~
\citep{feudal1992,SUTTON1999181,bacon2016optioncriticarchitecture,vezhnevets2017feudal}. The challenge is to generate, verify, and refine such skills while minimizing human engineering. LLMs can propose skill specifications including preconditions, effects, and reward functions, but these proposals are hypotheses that may be incorrect.

Prior work has LLMs generate reward functions or subgoal sequences that guide RL training~
\citep{read_and_reward,xie2023text2reward,ma2023eureka,sun2024large,liu2024rl}, but these approaches typically operate in a one-shot manner: the LLM proposes, and the RL agent executes, with no channel for execution feedback to correct the planner's assumptions. If the LLM overestimates resource needs or misses a hidden prerequisite, training fails and the error is never diagnosed. Effective LLM-RL integration requires a closed loop in which predictions are tested against execution and revised based on actual environment interaction.

We introduce Self-Supervised Composition and Learning of Skills with LLM Planning and Reinforcement Learning (\textsc{SCALAR}), a framework that couples symbolic LLM planning with low-level RL through a learned skill library (Figure~\ref{fig:scalar_diamond}). The LLM proposes skills with preconditions, effects, and reward functions; RL trains each skill and feeds back execution results to refine the LLM's specifications. Section~\ref{sec:preliminaries} formalizes skills as symbolic operators and neural options, Section~\ref{sec:scalar} details the LLM-RL loop, and Section~\ref{sec:experiments} evaluates on Craftax~
\citep{matthews2024craftax}, where \textsc{SCALAR} achieves 1.9$\times$ higher diamond collection than the best baseline on Craftax-Classic and 9\% success reaching the Gnomish Mines versus 0\% for prior methods. Our contributions are:
\begin{itemize}[leftmargin=*, itemsep=0pt, topsep=0pt]
    \item A bidirectional LLM-RL framework where trajectory feedback refines skill specifications
    \item \textbf{Pivotal Trajectory Analysis} for correcting preconditions and effects from successful executions
    \item \textbf{Frontier Checkpointing} to improve sample efficiency by reducing re-execution of trained skills
    \item Substantial improvements over prior methods on deep prerequisite tasks: 1.9$\times$ better diamond collection on Craftax-Classic and first reported success reaching the Gnomish Mines in full Craftax
\end{itemize}

\section{Related Work}

\textbf{Unsupervised Skill Discovery.}
Long-horizon tasks benefit from learning collections of skills rather than monolithic policies. Traditional methods discover behavioral primitives through exploration and graph-based representations \citep{bagaria2021skill,ecoffet2019go,evans2023creating}. Substantial research frames skill emergence as maximizing mutual information between states and skill labels \citep{vic,diayn,dads,hansen2019fast,liu2021aps,cic}, though discriminator-based objectives can saturate, yielding behaviors that differ only in subtle ways \citep{diayn}. Recent methods replace MI with Wasserstein dependency measures \citep{ozair2019wasserstein,he2022wasserstein,lsd,csd,metra}, pairing them with task-relevant state-space metrics. These methods discover skills without external guidance; \ourmethod{} instead leverages LLM priors to propose task-relevant skill structures directly.

\textbf{LLM-Generated Reward Functions.}
A natural starting point for LLM-RL integration is using LLMs to construct reward functions from task descriptions. Early work generates reward functions for complete tasks without direct agent involvement \citep{read_and_reward,xie2023text2reward,ma2023eureka}. Extensions enable dynamic reward adjustments based on agent interactions \citep{deng2025reward} or integrate LLM-generated subgoal hints into model rollouts \citep{liu2024worldmodelshintslarge}. These methods use LLMs for reward specification; \ourmethod{} goes further by having LLMs specify full skill structures including preconditions, postconditions, and compositional dependencies.

\textbf{LLM-Guided Skill Specification.}
Beyond reward functions, LLMs can specify complete skill structures. Methods in this space transform high-level goals into skill definitions with dense rewards and termination conditions \citep{leagueorginal,mao2025skill}, or generate subgoal sequences before training \citep{shek2025optiondiscoveryusingllmguided}. MaestroMotif \citep{klissarov2025maestromotif} combines human-authored skill descriptions with learned reward models for hierarchical skill learning. These approaches follow one-shot specification: the LLM proposes skill structures before training, with no mechanism to refine them based on execution outcomes.

\textbf{Symbolic Planning and RL.}
The skill structures we generate connect to classical AI planning. Symbolic operators with preconditions and effects enable STRIPS-style planning toward goals \citep{fikes1971strips}. Prior work integrates planning with hierarchical RL: SDRL \citep{sdrl2019} uses symbolic knowledge to guide subtask scheduling, PEORL \citep{peorl2018} uses hand-authored action models to guide option execution, and taskable RL \citep{illanes2020symbolic} uses symbolic plans as instructional scaffolding. These methods assume fixed, high-fidelity symbolic models. More recently, SayCan \citep{saycan} uses LLMs to propose high-level plans grounded through learned affordance models, and League++ \citep{li2024league} has an LLM propose plans verified by an A* planner. \ourmethod{} combines these threads: we use LLMs to specify STRIPS-like operators that classical planners sequence, but unlike prior work, we refine these specifications through trajectory analysis during training rather than assuming they are correct from the start.

\textbf{Refining LLM Priors through Interaction.}
When LLM priors prove insufficient, interaction can refine them. Prior work focuses on either reward functions \citep{deng2025reward} or agentic settings where LLMs execute code directly \citep{wang2023voyager,wu2024agentkit}. Voyager refines skills, but those skills are code-as-policy, not deep RL. \ourmethod{} addresses a different gap: refining symbolic skill specifications (preconditions, effects, resource counts) that guide deep RL training, combining LLM knowledge with the grounding provided by policy execution.

\section{Preliminaries}
\label{sec:preliminaries}

\subsection{Reinforcement Learning and Abstraction}

Consider a Markov Decision Process with state space $\mathcal{S}$, action space $\mathcal{A}$, transition dynamics $p(s' \mid s, a)$, and reward function $r: \mathcal{S} \times \mathcal{A} \to \mathbb{R}$. An agent follows a policy $\pi: \mathcal{S} \to \mathcal{A}$ to maximize cumulative discounted reward $\mathbb{E}[\sum_{t=0}^\infty \gamma^t r(s_t, a_t)]$.

For many complex tasks, reward functions are often defined over high-level task properties rather than raw observations. Achievement criteria such as ``craft an iron pickaxe'' are naturally expressed in terms of inventory contents and object placements. This implicitly assumes an \emph{abstraction function} $\phi: \mathcal{S} \to \mathcal{Z}$ mapping environment states to \emph{abstract states} $z \in \mathcal{Z}$:
\[
z = \{\texttt{has}(\texttt{wood}, 5),\; \texttt{placed}(\texttt{table}),\; \ldots\}
\]
Predicates like \texttt{has} and \texttt{placed} capture task-relevant properties. The same abstraction that enables symbolic reward definition can also support symbolic skill specification: skills are defined by preconditions and effects expressed as predicates over $\mathcal{Z}$.

\subsection{Skills as Operators and Options}

We formalize skills at two levels: \emph{operators} define what a skill achieves in terms of abstract state transitions, while \emph{options} implement the actual control policy.

\textbf{Operators.}
An \emph{operator} $o$ specifies how executing a skill transforms the agent's abstract state:
\[
o = \langle \textsc{Pre}(o),\; \textsc{Eff}^+(o),\; \textsc{Eff}^-(o) \rangle
\]
\emph{Preconditions} $\textsc{Pre}(o) \subseteq \mathcal{Z}$ are predicates required before execution. \emph{Positive effects} $\textsc{Eff}^+(o)$ hold upon success; \emph{negative effects} $\textsc{Eff}^-(o)$ are cleared. For example, crafting an iron pickaxe:

\begin{skillbox}[fontupper=\scriptsize,fonttitle=\bfseries\scriptsize]{Make Iron Pickaxe}
\setlength{\tabcolsep}{3.5pt}
\begingroup\setlength{\jot}{1pt}
\begin{tabular}{@{}l l@{}}
\textbf{Pre:} &
\(
\begin{aligned}[t]
& \texttt{placed}(\texttt{table}) \land \texttt{placed}(\texttt{furnace})\\
& \land \texttt{has}(\texttt{stone\_pickaxe}) \land\ \texttt{has}(\texttt{wood}, k_w) \land \texttt{has}(\texttt{iron}, k_i)\\
& \land \texttt{has}(\texttt{coal}, k_c) \land\ k_w \ge 2 \land k_i \ge 1 \land k_c \ge 1
\end{aligned}
\)\\[4pt]
\textbf{Eff$^{-}$:} & \( \texttt{has}(\texttt{wood}, 1),\ \texttt{has}(\texttt{iron}, 1),\ \texttt{has}(\texttt{coal}, 1) \)\\[2pt]
\textbf{Eff$^{+}$:} & \( \texttt{has}(\texttt{iron\_pickaxe}) \)
\end{tabular}
\endgroup
\end{skillbox}

\noindent This operator requires a crafting table, furnace, stone pickaxe, and sufficient raw materials. Upon success, it consumes wood, iron, and coal while producing an iron pickaxe.

\textbf{Options.}
Operators specify abstract transitions, but executing them requires taking actions in the MDP. An \emph{option} \citep{SUTTON1999181} provides this mapping through a learned policy $\pi_o$ and a termination function $\beta_o: \mathcal{Z} \to [0,1]$. Given state $s$ where $\textsc{Pre}(o) \subseteq \phi(s)$, the option executes $\pi_o$ until $\beta_o$ signals termination at $s'$ where $\textsc{Eff}^+(o) \subseteq \phi(s')$. Training requires a reward function $r_o: \mathcal{Z} \times \mathcal{A} \to \mathbb{R}$ that provides feedback toward achieving $\textsc{Eff}^+(o)$.

\textbf{Planning with Operators} The effects of one operator may satisfy the preconditions of another, inducing a \emph{dependency graph}. Given a goal, STRIPS-style planning \citep{fikes1971strips} computes a sequence of operators; executing the corresponding options in order transforms the environment to satisfy preconditions for subsequent skills. Option termination verifies that $\textsc{Eff}^+$ holds, but $\textsc{Eff}^-$ is only assumed correct since rewards do not constrain resource consumption.

\textbf{Training Options.} Options may be trained in isolation or compositionally. Isolated training optimizes each policy toward its own termination reward, which only verifies $\textsc{Eff}^+$; adverse $\textsc{Eff}^-$ such as over-consuming resources or leaving the environment in a poor state is not directly penalized. Compositional training executes options sequentially, computing advantages across the full trajectory:
\[
\hat{A}_t = \sum_{k=t}^{\infty} (\gamma\lambda)^{k-t} \delta_k
\]
These advantages extend beyond $\pi_i$'s termination. If $\pi_i$ over-consumes or damages the environment, downstream options suffer and returns propagate back, penalizing the adverse behavior. Isolated advantages truncate at termination, optimizing only for local success.

\newpage
\section{SCALAR: Self-Supervised Composition and Learning of Skills}
\label{sec:scalar}

\textsc{Scalar} learns a library of composable skills by leveraging an LLM to propose operators and reinforcement learning to train the corresponding options. Algorithm~\ref{alg:scalar-main} summarizes the approach: given a task description, we propose candidate operators, validate them, then iteratively train options for each skill needed to achieve the goal.

\subsection{Operator Proposal}
\label{sec:scalar:proposal}

Given a task description or game tutorial, an LLM proposes a set of candidate skills. This proceeds in stages: first proposing what each skill achieves, then grounding those specifications in executable code.

\textbf{Effects.} The LLM first proposes a set of skill names with their effects: what becomes true ($\textsc{Eff}^+$) and what is consumed ($\textsc{Eff}^-$) upon success. For a crafting game, this might yield skills like \textsc{CollectWood}, \textsc{PlaceTable}, \textsc{MineStone}, and \textsc{MakeStonePickaxe}. Each effect references predicates over the abstract state, such as $\texttt{has}(\texttt{pickaxe}, 2)$ indicating a stone-tier pickaxe.

\textbf{Preconditions.} Next, preconditions are defined in terms of effects from other skills. For \textsc{MakeStonePickaxe}, the LLM might specify:
\begin{center}
\small
$\textsc{Pre}: \texttt{has}(\texttt{wood}, 2) \land \texttt{has}(\texttt{stone}, 3) \land \texttt{near}(\texttt{table})$
\end{center}
These reference effects of other skills inducing the dependency structure that enables planning.

\textbf{Validation.} We filter proposed skills for \emph{feasibility} and \emph{novelty}. A skill is feasible if its preconditions are reachable from the initial state via some composition of already-validated skills. A skill is novel if its effects are not already achievable by existing skills. Skills failing either check are discarded.

\textbf{Termination and reward functions.} The LLM translates abstract specifications into executable code. The \emph{termination function} $\tau_o: \mathcal{Z} \to \{0, 1\}$ maps symbolic positive effects to a predicate over the environment state, and the \emph{reward function} $r_o: \mathcal{Z} \times \mathcal{A} \times \mathcal{Z}' \to \mathbb{R}$ provides learning signal. The simplest reward uses termination as a sparse signal: $r_o = \mathbf{1}[\tau_o(z') = 1]$. Optionally, the LLM generates dense shaping rewards to guide exploration:
\begin{lstlisting}[language=Python]
def done(z):
    return z.inv.pickaxe >= 2  # stone tier

def reward(z, z'):
    sparse = z'.inv.pickaxe - z.inv.pickaxe
    d_prev = norm(z.closest[TABLE])
    d_curr = norm(z'.closest[TABLE])
    dense = 0.01 * (d_prev - d_curr)
    return sparse + dense
\end{lstlisting}
The dense component rewards moving toward a crafting table, accelerating learning before the sparse crafting reward is achieved.

These proposals are \emph{hypotheses}: preconditions and effects may be incorrect or incomplete. Surviving hypotheses are refined through trajectory analysis during training (\S\ref{sec:scalar:trajectory}).

\begin{algorithm}[t]
\caption{\textsc{Scalar}: Self-Supervised Skill Learning}
\label{alg:scalar-main}
\begin{algorithmic}
\STATE \textbf{Input:} Task description $T$, goal $g$
\STATE \textbf{Output:} Trained skill library $\{(o, \pi_o)\}$
\STATE
\STATE $O \gets \textsc{LLM-Propose}(T)$ \hfill $\triangleright$ Propose operators
\STATE $O \gets \textsc{Validate}(O)$ \hfill $\triangleright$ Filter feasibility/novelty
\STATE $\Pi \gets \emptyset$
\WHILE{goal $g$ not achieved}
  \STATE $P \gets \textsc{Plan}(O, g)$ \hfill $\triangleright$ STRIPS plan
  \STATE $o^* \gets$ first untrained skill in $P$
  \STATE $P_{\mathrm{prereq}} \gets \textsc{Plan}(O, \textsc{Pre}(o^*))$
  \STATE $\Pi_{\mathrm{prereq}} \gets \{\pi_o : o \in P_{\mathrm{prereq}}\}$
  \STATE $(o^*, \pi_{o^*}, \Pi'_{\mathrm{prereq}}, \textit{succ}) \gets \textsc{TrainSkill}(o^*, \Pi_{\mathrm{prereq}})$
  \STATE $\Pi \gets (\Pi \setminus \Pi_{\mathrm{prereq}}) \cup \Pi'_{\mathrm{prereq}}$ \hfill $\triangleright$ Update prereqs
  \IF{$\textit{succ}$}
    \STATE $\Pi \gets \Pi \cup \{\pi_{o^*}\}$
  \ELSE
    \STATE Update $O$ with refined $o^*$
  \ENDIF
\ENDWHILE
\STATE \textbf{return} $\{(o, \pi_o) : o \in O, \pi_o \in \Pi\}$
\end{algorithmic}
\end{algorithm}

\subsection{Skill Composition}
\label{sec:scalar:composition}

Operators induce a dependency graph where one skill's effects satisfy another's preconditions. To train a new skill, we must first reach a state where its preconditions hold. A STRIPS planner computes a sequence of already-trained skills that achieves this. For \textsc{MakeStonePickaxe}:
\begin{center}
\small
\textsc{GetWood}(7) $\to$ \textsc{PlaceTable} $\to$ \textsc{MakeWoodPickaxe} $\to$ \textsc{GetStone}(3)
\end{center}
Mining stone requires a pickaxe, so we must first craft one. The planner accounts for resource consumption: 7 wood covers the table (4), wood pickaxe (3), and stone pickaxe (2), with the 2 leftover after crafting the wood pickaxe.

\subsection{Policy Training}
\label{sec:scalar:training}

Given a target skill to learn, training proceeds through rollout collection, advantage estimation, and policy updates. Algorithm~\ref{alg:train-skill} details the inner loop.

\textbf{Rollouts.} Each training iteration collects trajectories across $N$ parallel environments. Within each environment, we first execute the prerequisite options in order until preconditions hold, then execute the target option $\pi_{o^*}$ until termination .

\textbf{Frontier Checkpointing.} Training a target skill requires first executing its prerequisites to reach a state where preconditions hold. As the prerequisite chain deepens, an increasing fraction of training frames is spent re-executing already-learned skills rather than improving the target skill. Frontier Checkpointing addresses this by saving the environment state $s^*$ when preconditions first hold (the ``frontier'' between prerequisites and target), then restoring from $s^*$ with probability $\alpha_{\mathrm{reset}}$ on episode reset. Frontier Checkpointing assumes the ability to serialize and restore environment state, applicable in simulators or systems with state save/restore.

\textbf{Advantage estimation.} Each rollout produces trajectory segments for multiple policies. We compute advantages using Generalized Advantage Estimation (GAE):
\[
\hat{A}_t = \sum_{l=0}^{\infty} (\gamma \lambda)^l \delta_{t+l}, \quad \delta_t = r_t + \gamma V(s_{t+1}) - V(s_t)
\]
Crucially, the discounted returns extend across skill boundaries: an earlier skill's advantage includes future rewards from subsequent skills in the composition. This incentivizes each skill to terminate in states that help downstream skills succeed. The symbolic abstraction captures only part of the state. Predicates like $\texttt{has}(\texttt{iron})$ ignore health, enemy positions, and other factors. By propagating returns through the full composition, skills learn to respect these non-symbolic elements. For example, \textsc{CollectIron} learns not to rush toward ore if doing so leaves the agent at low health next to a hostile mob, even though the symbolic postcondition $\texttt{has}(\texttt{iron})$ would be satisfied.

\textbf{Policy updates.} We update all active policies on their respective trajectory segments using PPO. Rather than freezing prerequisite policies, joint updates prevent catastrophic forgetting. Without joint updates, prerequisite performance degrades as training shifts focus to later skills. While we use PPO, any on-policy algorithm (e.g., A2C, TRPO) would work; the key requirement is that all composed policies receive gradient updates from their execution segments.

\begin{algorithm}[t]
\caption{\textsc{TrainSkill}: Learn an Option for Operator $o^*$}
\label{alg:train-skill}
\begin{algorithmic}
\STATE \textbf{Input:} Target operator $o^*$, prerequisite options $\Pi_{\mathrm{prereq}}$
\STATE \textbf{Output:} Refined $o^*$, trained $\pi_{o^*}$, updated $\Pi_{\mathrm{prereq}}$, success flag
\STATE
\STATE Initialize $\pi_{o^*}$; \, $s^* \gets \texttt{None}$; \, $\textit{verified} \gets \texttt{False}$
\FOR{$\textit{iter} = 1$ \textbf{to} $\textit{max\_iter}$}
  \STATE \textit{// Rollout phase}
  \FOR{each step across $N$ parallel envs}
    \STATE Execute $\Pi_{\mathrm{prereq}}$ to reach $\textsc{Pre}(o^*)$
    \STATE $s^* \gets$ state where $\textsc{Pre}(o^*)$ first holds \hfill $\triangleright$ Checkpoint
    \STATE Execute $\pi_{o^*}$ toward $\textsc{Eff}^+(o^*)$
    \STATE On termination: restore $s^*$ w.p.\ $\alpha_{\mathrm{reset}}$
  \ENDFOR
  \STATE
  \STATE \textit{// Update phase}
  \STATE Compute GAE advantages for all transitions
  \STATE Update $\pi_{o^*}$ and all $\pi \in \Pi_{\mathrm{prereq}}$ via PPO
  \STATE
  \STATE \textit{// Analysis phase}
  \IF{$\neg\textit{verified}$ \textbf{and} any successful trajectories}
    \STATE $o'^* \gets \textsc{LLM-Refine}(o^*, \mathcal{T}_{\mathrm{success}})$ \hfill $\triangleright$ Verify specs
    \IF{$o'^* \neq o^*$}
      \STATE \textbf{return} $(o'^*, \emptyset, \Pi_{\mathrm{prereq}}, \texttt{False})$
    \ENDIF
    \STATE $\textit{verified} \gets \texttt{True}$
  \ENDIF
  \IF{$\textsc{SuccessRate} \geq \alpha_{\mathrm{grad}}$}
    \STATE \textbf{return} $(o^*, \pi_{o^*}, \Pi_{\mathrm{prereq}}, \texttt{True})$
  \ENDIF
\ENDFOR
\STATE \textit{// Training failed; refine from failures}
\STATE $o'^* \gets \textsc{LLM-Refine}(o^*, \mathcal{T}_{\mathrm{fail}})$
\STATE \textbf{return} $(o'^*, \emptyset, \Pi_{\mathrm{prereq}}, \texttt{False})$
\end{algorithmic}
\end{algorithm}

\subsection{Trajectory Analysis}
\label{sec:scalar:trajectory}
\begin{table*}[!htbp]
\centering
\small
\setlength{\tabcolsep}{4pt}
\renewcommand{\arraystretch}{1.10}
\begin{tabular}{@{}lccccc@{}}
\toprule
& \multicolumn{2}{c}{Craftax-Classic (\%)} & \multicolumn{3}{c}{Craftax (\%)} \\
\cmidrule(lr){2-3}
\cmidrule(lr){4-6}
Method & Eat Plant & Collect Diamond & Enter Dungeon & Unlock Gnomish Ladder & Enter Gnomish Mines \\
\midrule
\multicolumn{6}{l}{\textbf{Baselines}} \\
\addlinespace[1pt]
PPO-FC & 23.9 & 35.4 & 86.6 & 1.8 & 0.0 \\
PPO-RND & 21.9 & 34.5 & 87.4 & 1.9 & 0.0 \\
PPO-RNN & 0.0 & 40.9 & 89.7 & 0.0 & 0.0 \\
PPO-TRXL & 0.0 & 31.5 & \textbf{96.5} & 0.4 & 0.0 \\
PQN & 19.0 & 0.0 & 77.4 & 0.8 & 0.0 \\
PQN-RNN & 55.3 & 46.9 & \textbf{92.5} & 0.0 & 0.0 \\
\addlinespace[2pt]
\midrule
\multicolumn{6}{l}{\textbf{Our Method}} \\
\addlinespace[1pt]
SCALAR & \textbf{91.7} & \textbf{88.2} & 58.4 & \textbf{35.0} & \textbf{9.1} \\
w/o Frontier Checkpointing & \textbf{90.7} & \textbf{88.2} & 58.9 & 11.5 & 1.3 \\
+w/o Trajectory Analysis & 0.0 & 67.3 & 22.9 & 0.4 & 0.0 \\
SCALAR-TRXL & 0.0 & 45.0 & 85.2 & \textbf{34.1} & 8.2 \\
\bottomrule
\end{tabular}
\caption{Success rates on Eat Plant and Collect Diamond, the two most difficult long-horizon tasks in Craftax-Classic, and Enter Dungeon, Unlock Gnomish Ladder, and Enter Gnomish Mines, the three deepest achievements in Craftax. Results at 1B frames, averaged over 5 seeds. Ablation rows show cumulative removal of components.}
\label{tab:results}
\vspace{-1em}
\end{table*}

Trajectory analysis triggers when training achieves nonzero success rate. We analyze the first successful trajectories to verify and refine the operator specification. For each successful episode, we record:
\[
(z_{\mathrm{start}},\; z_0 \to z_1 \to \cdots \to z_T,\; z_{\mathrm{end}})
\]
the abstract state when the option initiated, the complete trajectory of intermediate states, and the final state at termination. We prompt the LLM with the hypothesized operator alongside these observations:
\[
o' \gets \textsc{LLM-Refine}(o,\; \{(z_{\mathrm{start}}, z_{\mathrm{end}}, \{z_t\})\}_{\mathrm{success}})
\]
The LLM compares them and outputs a revised specification: preconditions updated to reflect what was actually present at $z_{\mathrm{start}}$, positive effects updated to match what was achieved at $z_{\mathrm{end}}$, and negative effects updated to reflect consumption.

Recording only endpoints is insufficient because the policy may compensate for missing preconditions mid-episode. Suppose \textsc{MakeIronPickaxe} requires 3 iron, but the planner delivers only 2. In rare episodes where iron happens to be nearby, the policy collects it and succeeds. Endpoints alone would suggest 2 iron suffices, but most episodes fail. Full trajectories reveal the policy collected additional iron, exposing the true requirement.

If the refined operator differs from the current one, we discard the partially-trained policy and restart with the corrected specification. This catches errors early rather than discovering them when composing skills later.

\section{Experiments}
\label{sec:experiments}

We evaluate \textsc{Scalar} on Craftax-Classic and Craftax, sparse-reward survival environments where tasks decompose into prerequisite chains. Our central claim is that LLM-specified compositional skills outperform monolithic RL on long-horizon sparse tasks. Section~\ref{sec:main_results} validates this claim by comparing SCALAR against strong baselines. We then dissect the mechanisms behind this improvement through three ablations: trajectory analysis (Section~\ref{sec:ablation_pta}), which corrects LLM priors from execution feedback; Frontier Checkpointing (Section~\ref{sec:ablation_reset}), which improves sample efficiency on deep prerequisite chains; and online adaptation (Section~\ref{sec:ablation_ood}), which enables recovery from specification errors under distribution shift.

\begin{figure*}[t]
  \centering
  \includegraphics[width=\textwidth]{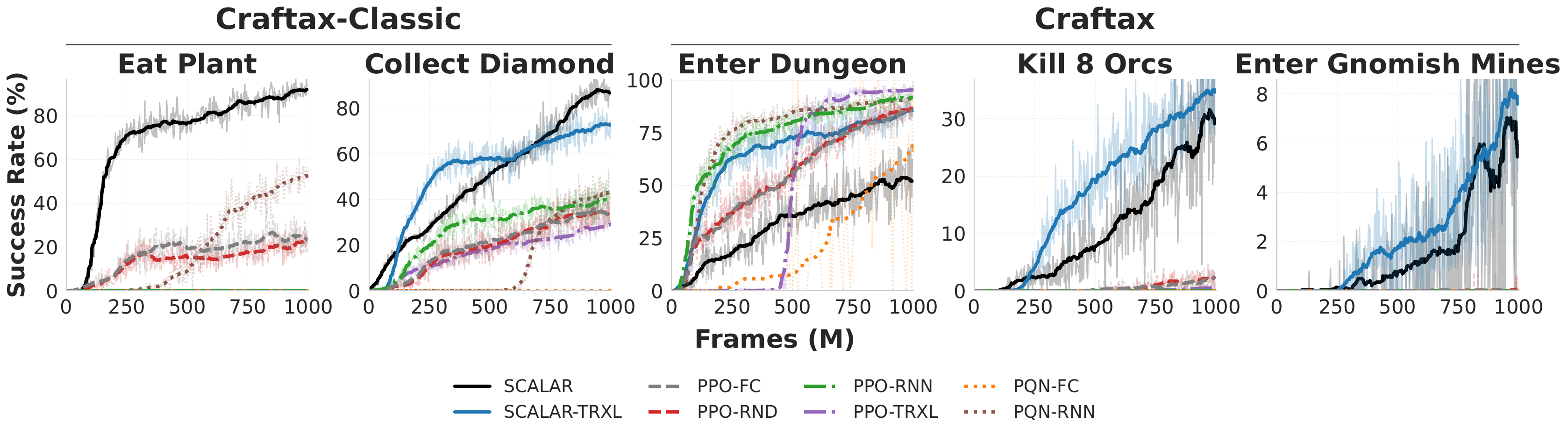}
  \vskip -0.5em
  \caption{Training curves on Craftax-Classic (left two panels) and Craftax (right three panels). X-axis: training frames in millions. Y-axis: success rate (\%) over 256 evaluation episodes. Lines show mean over 5 seeds; shaded regions show standard error. SCALAR (black solid) and SCALAR-TRXL (blue solid) are our method; dashed lines are PPO variants; dotted lines are PQN variants.}
  \label{fig:main_results}
  \vskip -0.8em
\end{figure*}

\subsection{Experimental Setup}
\label{sec:exp_setup}

\textbf{Environments.} Craftax \citep{matthews2024craftax} is an open-ended RL benchmark inspired by Crafter \citep{crafter} and Nethack \citep{kuttler2020nle}. We evaluate on two variants. \textit{Craftax-Classic} features a single $64 \times 64$ overworld where agents gather resources, craft tools, and mine diamond through a 10+ step prerequisite chain. Success requires maintaining health via sleep, food, and water. Diamonds spawn at least once per map, and the 10,000-step limit rewards methods that balance survival with exploration. \textit{Craftax} extends this to 9 procedurally generated floors. The overworld shrinks to $48 \times 48$ but resources grow scarcer: iron is $1.5\times$ rarer and diamond is $5\times$ rarer with no spawn guarantee. Combat becomes mandatory for progress. Entering the Gnomish Mines, for instance, requires killing 8 orcs. See Appendix~\ref{app:environments} for detailed specifications.

\textbf{Baselines.} All experiments use end-to-end JAX RL with vectorized environments. Static computation graphs and JIT compilation enable high throughput: 1B frames in approximately one hour on a single H100. This speed comes with constraints on which methods can be compared. We use PPO and PQN, which have mature JAX implementations. We evaluate four architectures: feedforward (FC), recurrent (RNN), Transformer-XL (TRXL), and RND exploration bonus. All methods observe a symbolic state including inventory counts, a 7$\times$9 local map, and relative positions to key entities (Appendix~\ref{sec:app-encoder}). We train for 1B frames with 5 seeds.

\textbf{SCALAR.} SCALAR uses the same architectures as baselines but with a mixture-of-experts structure. Each skill maintains separate actor-critic heads while sharing a common backbone: FC shares the first layer with 3 skill-specific layers per expert; TRXL shares the entire transformer backbone with 2 skill-specific linear layers per expert. SCALAR-FC uses Frontier Checkpointing to save and restore environment states, skipping prerequisite execution during training. SCALAR-TRXL forgoes checkpointing because restoring environment state also requires restoring the transformer's memory buffer, which is difficult under static JAX compilation (Appendix~\ref{sec:app-compute}). Skills graduate at 30\% success; goal tasks train to 99\%. See Appendix~\ref{app:exp_setup} for implementation details, Appendix~\ref{sec:app-hyperparams} for hyperparameters, and Appendix~\ref{sec:app-graduation-threshold} for an ablation of the graduation threshold.

\subsection{Comparison with Baselines}
\label{sec:main_results}

Table~\ref{tab:results} shows final success rates; Figure~\ref{fig:main_results} plots training curves. On Craftax-Classic, SCALAR-FC achieves 88.2\% diamond collection versus 46.9\% for the best baseline (PQN-RNN), a 1.9$\times$ improvement (2.5$\times$ over the best PPO baseline at 35.4\%). The gap reflects focused versus diffuse learning: baselines receive reward for all achievements equally, spreading signal across tasks like sword crafting that are irrelevant to diamond, while SCALAR trains only the skills needed for the target.

On intermediate tasks like Enter Dungeon, SCALAR lags baselines (58\% for FC, 85\% for TRXL versus 90--96\% for baselines). This gap exposes a tradeoff in goal-directed learning. SCALAR skips combat with overworld zombies and skeletons because they are not prerequisites for dungeon entry. An agent can reach the dungeon without killing them, so trajectory analysis never identifies combat as required. Baselines, however, receive reward for all achievements and incidentally develop combat proficiency that transfers to dungeon survival. Focused learning concentrates training signal on the critical path but sacrifices indirect skill transfer from auxiliary behaviors (Appendix~\ref{sec:app-behavioral}).

This tradeoff favors SCALAR when tasks require structured prerequisite chains. Deep chains amplify SCALAR's advantage: each skill is trained to 30\% success before advancing, ensuring reliable execution. Baselines must learn all behaviors simultaneously, and the 8-orc combat requirement is difficult to master. PPO-TRXL achieves 96\% on Enter Dungeon but fails on Enter Gnomish Mines (0\%) because the 8-orc combat requirement blocks progress. Both SCALAR variants reach approximately 9\% by training separate skills for dungeon navigation, combat, and mine entry.

\subsection{Ablation: Trajectory Analysis}
\label{sec:ablation_pta}

Accurate skill specifications are essential for SCALAR's performance. Trajectory analysis corrects LLM priors by examining the first successful trajectories; without this feedback loop, SCALAR reduces to one-shot LLM skill proposal methods~\citep{leagueorginal,li2024league,mao2025skill,shek2025optiondiscoveryusingllmguided}. For Collect Diamond, the LLM initially proposes 12 operators, 8 of which pass feasibility and novelty filtering. Trajectory analysis triggers 3 refinement events that correct resource count estimates for wood, stone, and iron.

\begin{figure}[t]
  \centering
  \includegraphics[width=\columnwidth]{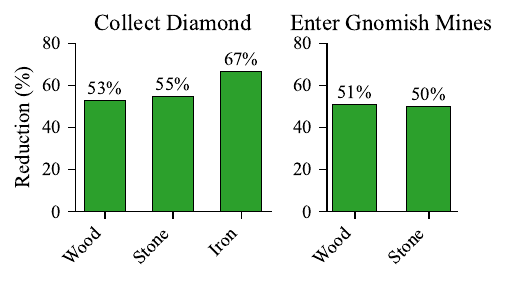}
  \vskip -0.5em
  \caption{Resources collected before attempting diamond, comparing SCALAR with and without trajectory analysis. Bars show total wood, stone, and iron gathered. Left bars: LLM's initial estimates. Right bars: after trajectory analysis refines requirements.}
  \label{fig:resource-collection}
  \vskip -1.8em
\end{figure}

The LLM's initial priors substantially overestimate resource requirements: 19 wood, 11 stone, and 3 iron before diamond, when only 9 wood, 5 stone, and 1 iron are needed. See Appendix~\ref{app:exp_setup} for trajectory analysis parameters and Appendix~\ref{sec:app-trajectory-analysis} for sensitivity analysis. Figure~\ref{fig:resource-collection} shows 50--67\% reductions in resource collection after trajectory analysis corrects these estimates. Over-collection wastes frames on prerequisite skills that could train the target skill instead. Table~\ref{tab:results} quantifies the performance impact: removing trajectory analysis drops diamond collection from 88\% to 67\%, and Enter Gnomish Mines from 9\% to 0\%.

Trajectory analysis also discovers prerequisites that LLMs cannot infer from documentation alone. Eat Plant achieves 92.6\% with trajectory analysis versus 0\% without. The task requires approximately 600 steps for plant maturation, during which the agent must maintain health through sleep, eating, and drinking. These survival behaviors are essential but absent from the game manual, so the LLM cannot infer them zero-shot. By observing successful episodes, trajectory analysis identifies what the agent actually needed to survive.

\subsection{Ablation: Frontier Checkpointing}
\label{sec:ablation_reset}

Accurate specifications alone do not guarantee learning. The agent must also spend sufficient frames training each skill. Deep prerequisite chains create a sample efficiency problem: if entering the dungeon takes approximately 500 steps, few frames remain for training combat or mine entry within a single episode. Frontier Checkpointing addresses this by resetting to saved states where prerequisites are satisfied.

\begin{figure}[t]
  \centering
  \includegraphics[width=\columnwidth]{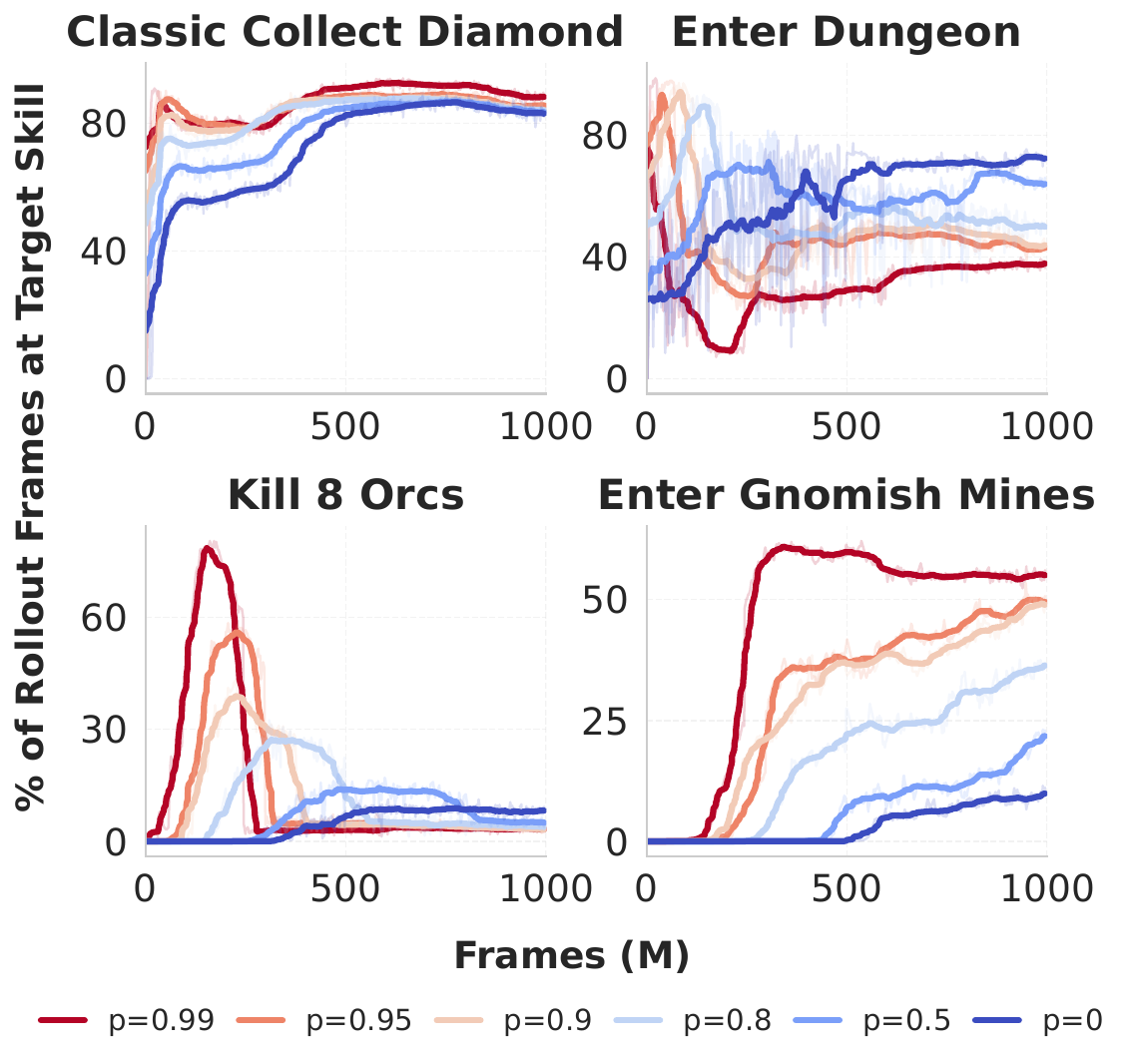}
  \vskip -0.5em
  \caption{Fraction of training frames spent on the target skill (y-axis) for different reset probabilities $p \in \{0, 0.5, 0.9, 0.99\}$. Four panels: Craftax-Classic (Collect Diamond) and Craftax (Enter Dungeon, Kill 8 Orcs, Enter Gnomish Mines). Higher $p$ means more frequent resets to saved frontier states.}
  \label{fig:reset-ablation}
  \vskip -1.0em
\end{figure}

Figure~\ref{fig:reset-ablation} shows how reset probability affects frame allocation. Without reset ($p=0$), nearly 0\% of frames train Kill 8 Orcs or Enter Gnomish Mines. With reset ($p \geq 0.9$), over 60\% of frames target the current skill. Even without Frontier Checkpointing, SCALAR-FC achieves 1.3\% on Enter Gnomish Mines versus 0\% for PPO-FC, while checkpointing improves this to 9.1\% (Table~\ref{tab:results}). SCALAR-TRXL reaches 8.2\% without any frontier reset; its memory-based exploration partially compensates for the frame allocation disadvantage. Extending Frontier Checkpointing to stateful architectures remains promising future work.

When prerequisites are short, Frontier Checkpointing provides marginal benefit. On Collect Diamond, all reset probabilities converge to 85--90\% final success because the prerequisite chain executes quickly, leaving sufficient frames for diamond training even without reset. See Appendix~\ref{sec:app-reset-ablation} for a comprehensive ablation including the diversity-overfitting tradeoff at extreme reset values.

\begin{table}[H]
\centering
\small
\caption{Diamond collection success (\%) under modified crafting recipes. Each row changes one recipe from the default (e.g., Iron Pickaxe requires 2 iron instead of 1).}
\label{tab:ood}
\begin{tabular}{@{}lccc@{}}
\toprule
Modification & PPO & \textsc{Scalar} & + Traj. \\
\midrule
Baseline & 34.8 & 88.1 & 88.1 \\
Iron Pick (2 Iron) & 5.1 & 5.4 & \textbf{86.1} \\
Iron Pick (2 Wood) & 14.8 & 61.0 & \textbf{81.8} \\
Iron Sword (3 Iron) & 3.1 & 88.1 & 87.0 \\
Table (3 Wood) & 32.8 & 0.7 & \textbf{86.7} \\
\bottomrule
\end{tabular}
\vspace{-1em}
\end{table}

\subsection{Ablation: Online Adaptation}
\label{sec:ablation_ood}

Trajectory analysis continues post-training. Examining execution trajectories lets it refine skill specifications after deployment when dynamics change. Table~\ref{tab:ood} tests this with modified crafting recipes. Each variant stresses a different aspect of adaptation. Increasing the iron pickaxe requirement from 1 to 2 iron tests whether trajectory analysis detects additional resource consumption; without correction, SCALAR drops to 5.4\%, matching PPO, because the LLM's specification remains wrong. With trajectory analysis, SCALAR recovers to over 86\% by observing that successful episodes collected more iron than originally specified. Increasing the crafting table cost from 2 to 3 wood is more challenging because it changes a skill early in the prerequisite chain; without correction, SCALAR drops to 0.7\%, yet trajectory analysis still recovers to 86.7\%. The Iron Sword variant shows robustness to distractor changes: even when an irrelevant recipe changes, trajectory analysis does not overcorrect. Trajectory analysis thus provides robustness to specification errors, not merely efficiency gains.

\section{Conclusion}

We introduce SCALAR, a method for learning composable skills in long-horizon, sparse-reward environments. SCALAR couples LLM planning with RL through a bidirectional feedback loop: the LLM proposes symbolic operators; RL trains policies and feeds back execution results to refine specifications. Trajectory analysis corrects LLM priors by examining successful trajectories, enabling adaptation to environment dynamics and recovery from specification errors. On Craftax and Craftax-Classic, SCALAR achieves state-of-the-art performance on deep prerequisite tasks: 88\% diamond collection versus 47\% for the best baseline, and 9\% on Enter Gnomish Mines where prior methods fail entirely. Trajectory analysis further enables online adaptation when environment dynamics shift, recovering from specification errors without manual intervention.

\textbf{Limitations.}
SCALAR's design imposes constraints. First, skills require symbolic state and a predefined vocabulary: preconditions and effects must be expressible within the state encoder, limiting application to settings where such structure is available. Second, the skill decomposition enforces fixed execution order, preventing opportunistic interleaving of order-invariant subtasks. Third, Frontier Checkpointing requires state serialization and introduces a diversity-efficiency tradeoff; extending to stateful models or real-world settings without native serialization remains future work. See Appendix~\ref{sec:app-limitations} for details.

\newpage

\section*{Impact Statement}

This work develops SCALAR, a framework that couples LLM planning with reinforcement learning through learned skills and symbolic operators. We discuss potential positive impacts, broader applications, risks, and limitations.

\textbf{Positive Societal Impacts.} SCALAR reduces the manual engineering burden for training RL agents on complex tasks. By leveraging LLM knowledge to propose skill structures and refining them through environment interaction, the framework makes long-horizon RL more accessible to practitioners without extensive domain expertise in reward engineering. The compositional nature of learned skills enables reusable behavior libraries that can be recomposed for new tasks without retraining from scratch. Frontier Checkpointing optionally improves sample efficiency, reducing computational costs and environmental impact compared to methods requiring longer training.

\textbf{Broader Applications.} Beyond game-playing benchmarks, SCALAR's approach may extend to robotic manipulation (where skills correspond to grasping, placing, and tool use), UI automation (sequential interactions with interface elements), and industrial process control (compositional procedures with verification steps). The bidirectional feedback between high-level planning and low-level execution provides a template for grounding LLM reasoning in physical environments.

\textbf{Risks and Mitigation.} More capable autonomous agents carry risks of misuse, including automation of tasks intended to require human judgment or oversight. Deploying such systems requires domain constraints, monitoring mechanisms, and careful evaluation before real-world application. SCALAR's reliance on LLM priors may propagate biases present in the language model; trajectory analysis mitigates this partially but does not eliminate it entirely.

\textbf{Limitations Affecting Deployment.} SCALAR requires a predefined symbolic abstraction, limiting direct application to settings where such structure is unavailable or difficult to specify. Frontier Checkpointing, while beneficial for sample efficiency, requires environment state serialization and is optional; SCALAR functions without it but trains more slowly on deep prerequisite chains. The fixed execution order enforced by skill decomposition prevents opportunistic interleaving of order-invariant subtasks. These constraints suggest SCALAR is most suitable for simulated environments and structured real-world settings where symbolic state is available.

\section*{Acknowledgements}

This material is based upon work supported by the National Science Foundation Graduate Research Fellowship Program under Grant No.\ DGE-2140739, the Defense Advanced Research Projects Agency (DARPA) under Contract No.\ FA8750-23-2-1015 (ANSR), and the Office of Naval Research under Grant No.\ N00014-23-1-2840. This work used DeltaAI at the University of Illinois Urbana-Champaign through allocation CIS240128 from the Advanced Cyberinfrastructure Coordination Ecosystem: Services \& Support (ACCESS) program, which is supported by U.S.\ National Science Foundation grants \#2138259, \#2138286, \#2138307, \#2137603, and \#2138296. This research was also supported with Cloud TPUs from Google's TPU Research Cloud (TRC). Any opinions, findings, and conclusions or recommendations expressed in this material are those of the author(s) and do not necessarily reflect the views of the National Science Foundation, DARPA, the Office of Naval Research, or the U.S.\ Government.

\bibliographystyle{icml2026}
\bibliography{references}

\newpage
\appendix
\onecolumn
\newpage
\section{Environment Details: Craftax-Classic vs Craftax}
\label{app:environments}

\begin{figure}[H]
  \centering
  \includegraphics[width=0.7\columnwidth]{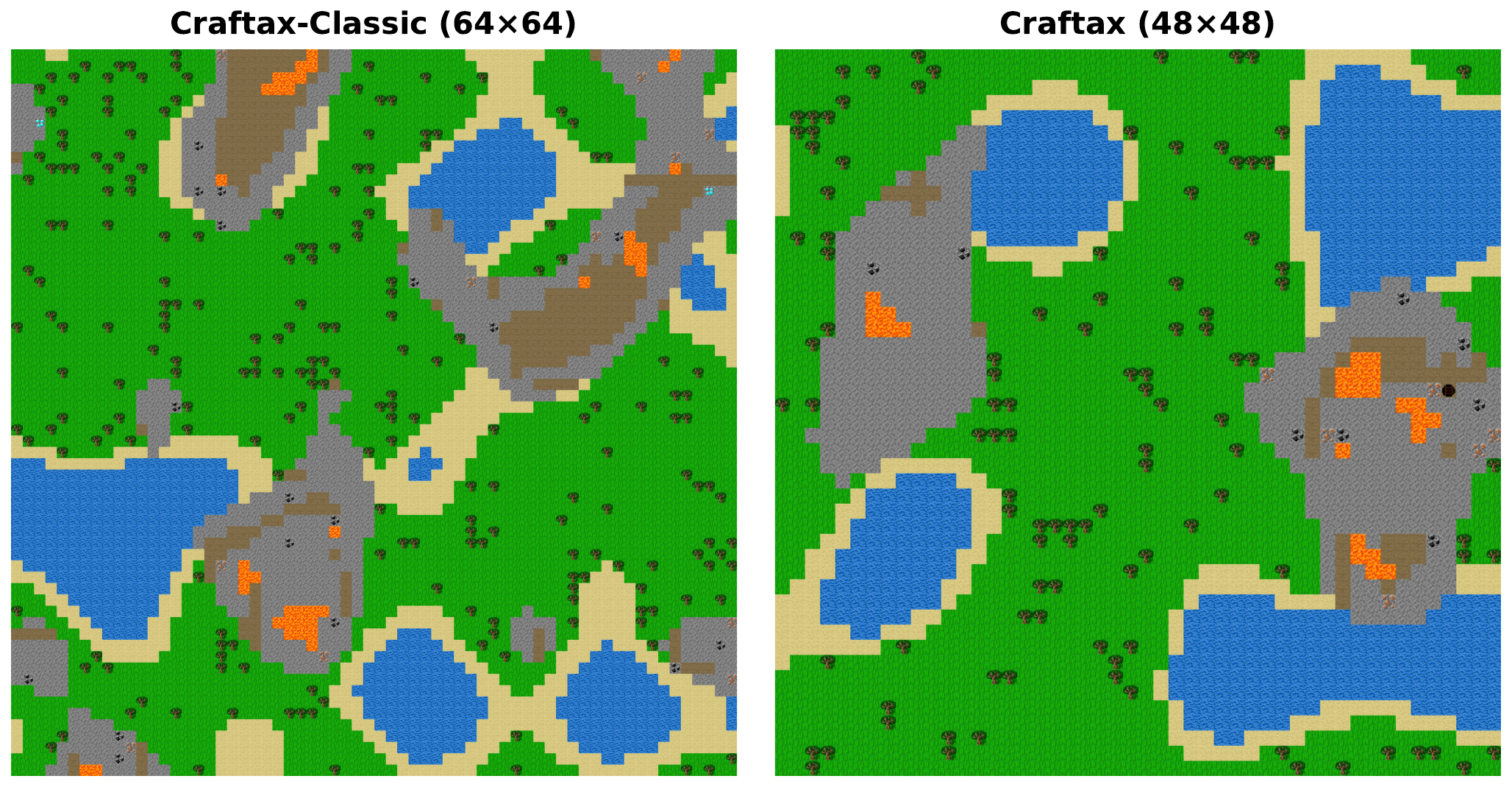}
  \caption{Comparison of Craftax-Classic and Craftax environments. Craftax-Classic is shown on the left; Craftax on the right.}
  \label{fig:world-comparison}
\end{figure}

Craftax-Classic and Craftax are two variants of the Craftax benchmark with different world structures and resource distributions. Figure~\ref{fig:world-comparison} shows both environments.

Craftax-Classic is a single-floor survival task on a $64 \times 64$ overworld. The agent must collect resources, craft tools, and mine diamond. At least one diamond is guaranteed per map, iron spawns at 3\% of stone tiles, and diamond at 0.5\%. The 10,000 timestep limit and guaranteed diamond make the task tractable for methods that can survive long enough to explore.

Craftax features 9 procedurally generated floors: an overworld plus 8 dungeon levels. The overworld is smaller at $48 \times 48$ with scarcer resources: iron is 1.5$\times$ rarer and diamond is 5$\times$ rarer than in Classic, with no spawn guarantee. Reaching deeper floors requires combat: entering the Gnomish Mines requires killing 8 orcs on the first dungeon floor. Diamond becomes more common in deeper floors at 0.5\% in Gnomish Mines and 1\% in Troll Mines, but reaching these floors requires combat proficiency. The action space expands from 17 to 43 actions, and the timestep limit increases to 100,000.

These differences explain some patterns in the results. On Craftax, most baselines achieve 80--95\% success on Enter Dungeon as shown in Table~\ref{tab:results} because the smaller overworld means random exploration often reaches the staircase before the agent dies. On Craftax-Classic, the same baselines achieve only 35--40\% on Collect Diamond despite the guaranteed spawn because the larger map requires sustained, directed exploration that outlasts typical survival times. The bottleneck shifts from finding the objective to surviving long enough to reach it.

\newpage
\section{Further Results}
\label{app:further_results}

\subsection{FC vs Transformer-XL Comparison}
\label{sec:app-architecture}

We compare SCALAR with fully-connected networks against a Transformer-XL backbone. The key difference is that FC uses Frontier Checkpointing with resetting to saved states during training, while TRXL does not. This is because restoring environment state would also require restoring the transformer's memory buffer, which is difficult to implement under static JAX compilation. See Appendix~\ref{sec:app-compute} for details. Figure~\ref{fig:fc-vs-trxl} shows the comparison across four tasks.

\begin{figure}[H]
  \centering
  \includegraphics[width=0.9\columnwidth]{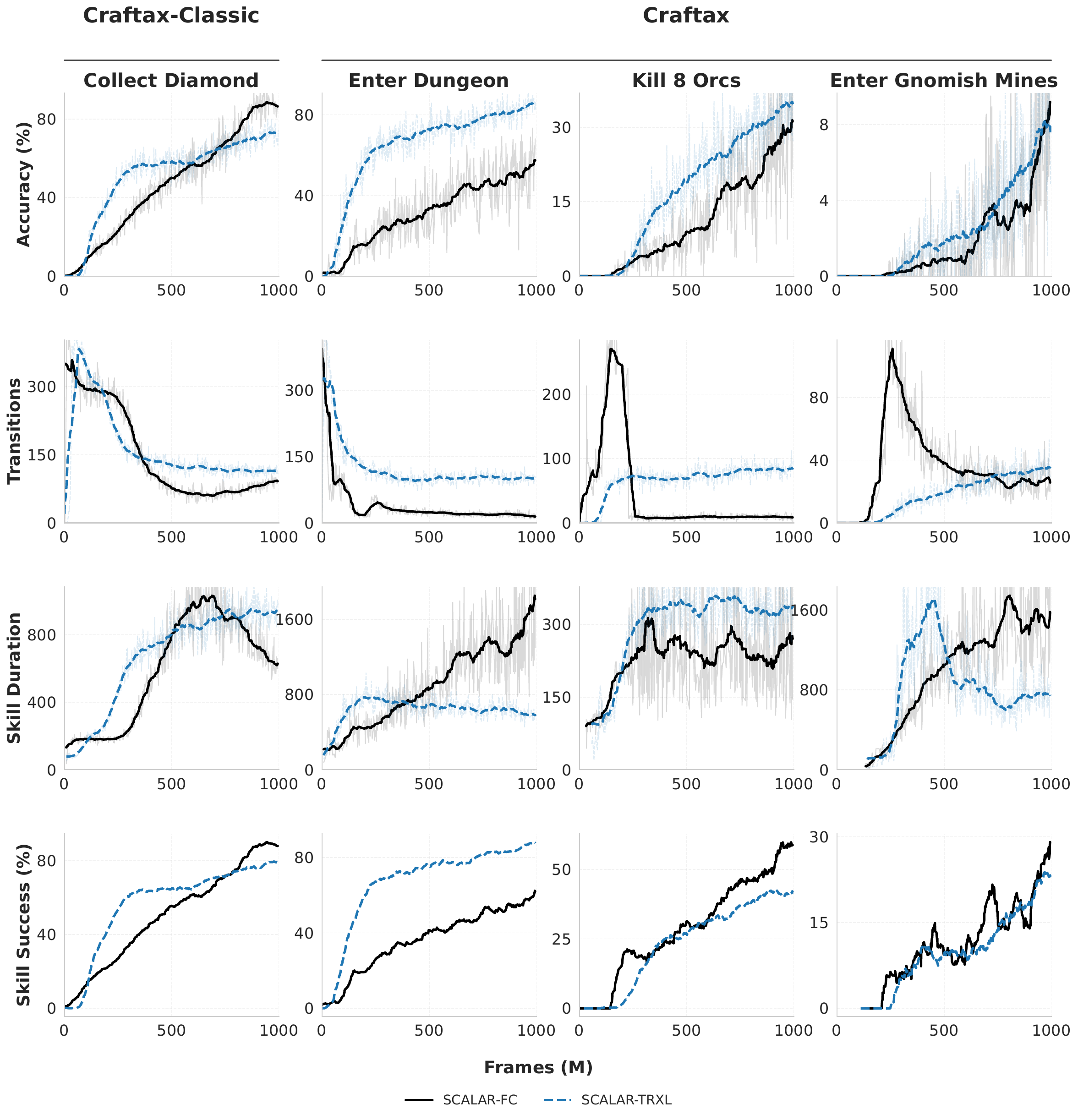}
  \caption{Comparison of SCALAR-FC and SCALAR-TRXL across four tasks. \textbf{Row 1:} Overall task success rate. \textbf{Row 2:} Number of skill terminations per training batch. \textbf{Row 3:} Frames spent at the target skill per episode. \textbf{Row 4:} Success rate conditioned on reaching the target skill.}
  \label{fig:fc-vs-trxl}
\end{figure}

\paragraph{TRXL explores more efficiently, but Frontier Checkpointing closes the gap.} TRXL is a larger architecture with memory over past observations. On Enter Dungeon, this enables more targeted exploration: TRXL achieves approximately 80\% accuracy versus approximately 60\% for FC, and Row 3 shows TRXL's skill duration peaks earlier, indicating it finds the dungeon entrance faster. Without memory, FC explores by fleeing from mobs, which works in large open maps but causes oscillation in narrow dungeon corridors.

This efficiency matters for training deeper skills. Reaching the Gnomish Mines requires first entering the dungeon; if dungeon entry takes many steps, fewer frames remain for training later skills. TRXL's faster dungeon navigation would give it an advantage on deeper skills, but FC's Frontier Checkpointing neutralizes this by skipping the dungeon traversal entirely during later skill training. The result is that despite architectural differences, both achieve similar final performance on Enter Gnomish Mines at approximately 9\% for FC and approximately 8\% for TRXL.

On Craftax-Classic with the Collect Diamond task, both architectures converge to similar performance. The larger overworld map suits FC's flee-based exploration, and the prerequisite chain is short enough that Frontier Checkpointing provides less benefit.

For Kill 8 Orcs, FC outperforms TRXL at approximately 35\% versus approximately 25\%. TRXL spends more frames at the target skill as shown in Row 3 yet achieves lower skill success. The transformer is more sensitive to health loss and prioritizes survival over aggressive combat, reducing kill efficiency against the 8-orc threshold.

\newpage
\subsection{Trajectory Analysis}
\label{sec:app-trajectory-analysis}

\begin{figure}[H]
  \centering
  \includegraphics[width=0.6\columnwidth]{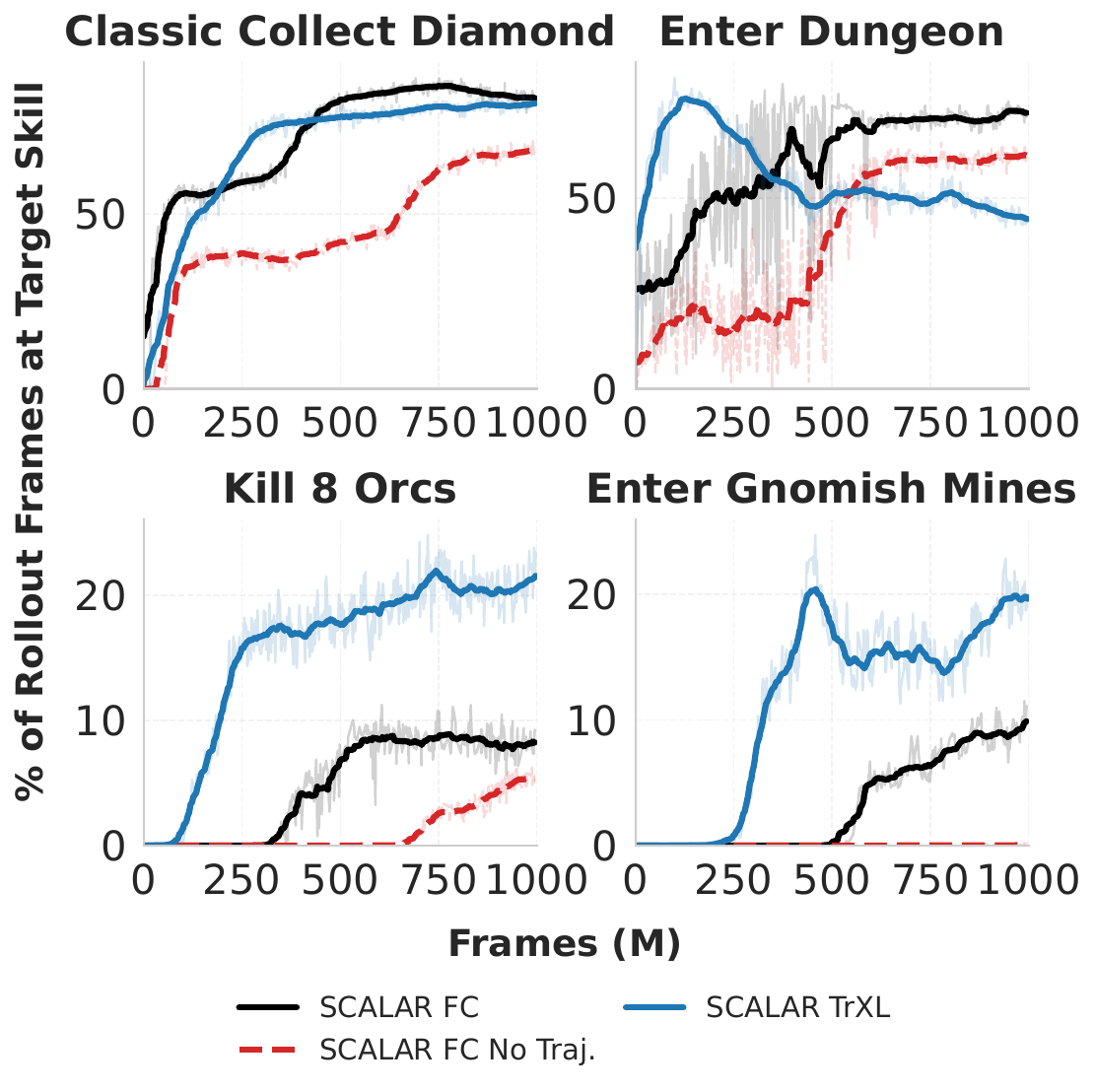}
  \vskip -0.5em
  \caption{Fraction of training frames spent on the target skill over training. Four panels show Craftax-Classic Collect Diamond and Craftax Enter Dungeon, Kill 8 Orcs, and Enter Gnomish Mines. Solid lines show results with trajectory analysis; dashed lines show results without trajectory analysis.}
  \label{fig:frontier-efficiency-appendix}
\end{figure}

Trajectory analysis refines LLM-proposed skill specifications by examining successful rollouts. Figure~\ref{fig:frontier-efficiency-appendix} shows the fraction of training frames spent on target skills with and without trajectory analysis.

Without trajectory analysis, the LLM's initial resource estimates are based on Minecraft-like priors, such as 3 iron for an iron pickaxe instead of 1. These overestimates cascade through the dependency graph: training spends excessive frames on resource collection skills before reaching the target. On Craftax-Classic, this reduces diamond training efficiency from approximately 75\% to approximately 50\% of frames. On full Craftax, the effect is more severe: without correction, near 0\% of frames reach Kill 8 Orcs or Enter Gnomish Mines, as the prerequisite chain consumes the entire episode.

Trajectory analysis also discovers non-obvious prerequisites. For Eat Plant, the LLM cannot infer from the game manual that plants require approximately 600 steps to mature, during which the agent must maintain health through sleep, food, and water. Without trajectory analysis, Eat Plant achieves 0\% success; with analysis, it achieves 92\%.

\paragraph{Sensitivity to number of trajectories.} We use $K=10$ successful trajectories for analysis. This provides sufficient diversity to detect systematic errors while keeping LLM context manageable. Results are robust to modest variation in $K$; the key requirement is observing enough successful executions to distinguish genuine requirements from incidental states.

\newpage
\subsection{Hyperparameter Ablations}
\label{sec:app-hyperparameters}

We ablate three key hyperparameters: reset probability, graduation threshold, and network sharing.

\subsubsection{Reset Probability}
\label{sec:app-reset-ablation}

\begin{figure}[H]
  \centering
  \includegraphics[width=0.9\textwidth]{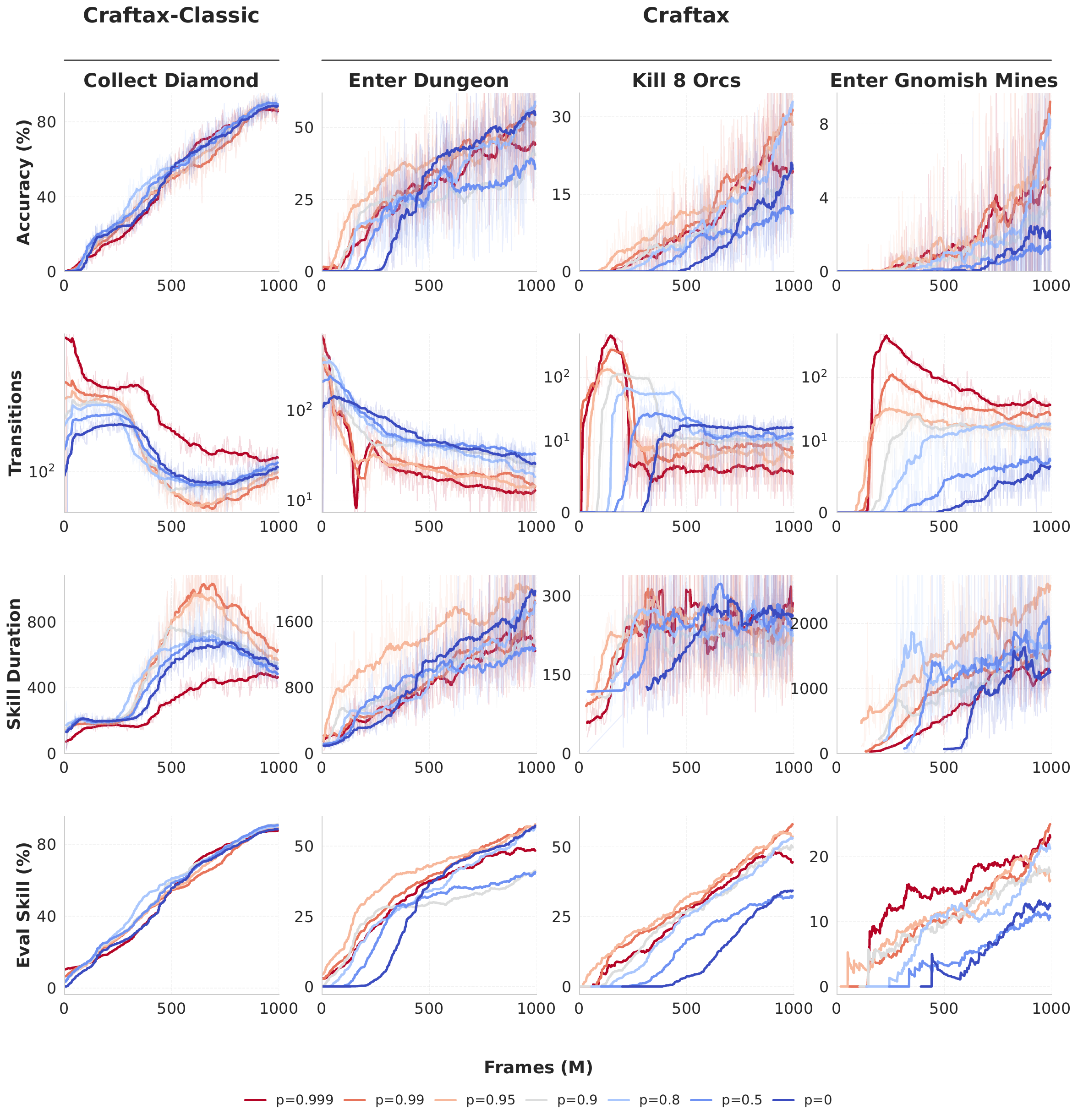}
  \caption{Effect of reset probability $\alpha_{\mathrm{reset}}$ on training across four tasks. \textbf{Row 1:} Overall success rate. \textbf{Row 2:} Success rate conditioned on reaching the target skill. \textbf{Row 3:} Frames per episode at the target skill. \textbf{Row 4:} Skill terminations per batch. Higher reset values are shown in red with $\alpha = 0.999$ and are essential for tasks with long prerequisites such as Kill 8 Orcs and Enter Gnomish Mines, while shorter tasks like Collect Diamond are insensitive to reset probability.}
  \label{fig:reset-comprehensive}
\end{figure}

We ablate the reset probability $\alpha_{\mathrm{reset}}$ from 0, meaning never reset to checkpoint, to 0.999, meaning almost always reset. Figure~\ref{fig:reset-comprehensive} shows the effect across four tasks.

\paragraph{Reset is essential when prerequisites dominate episode length.} For Kill 8 Orcs and Enter Gnomish Mines, high reset with $\alpha \geq 0.9$ achieves 30--40\% and 8--10\% success respectively, while $\alpha = 0$ remains near 0\% throughout training. Without reset, the agent spends all training time re-executing prerequisites. Entering the dungeon alone takes approximately 500 steps, leaving near-zero frames for the target skill. Row 3 confirms this: at $\alpha=0$, the agent spends less than 5\% of frames at the target skill for Enter Gnomish Mines, while $\alpha=0.99$ spends more than 60\%.

For Collect Diamond, all reset probabilities converge to approximately 85--90\%. The prerequisite chain including resource collection and crafting executes quickly enough that even $\alpha=0$ provides sufficient training signal for the target skill.

\begin{figure}[H]
  \centering
  \includegraphics[width=0.85\textwidth]{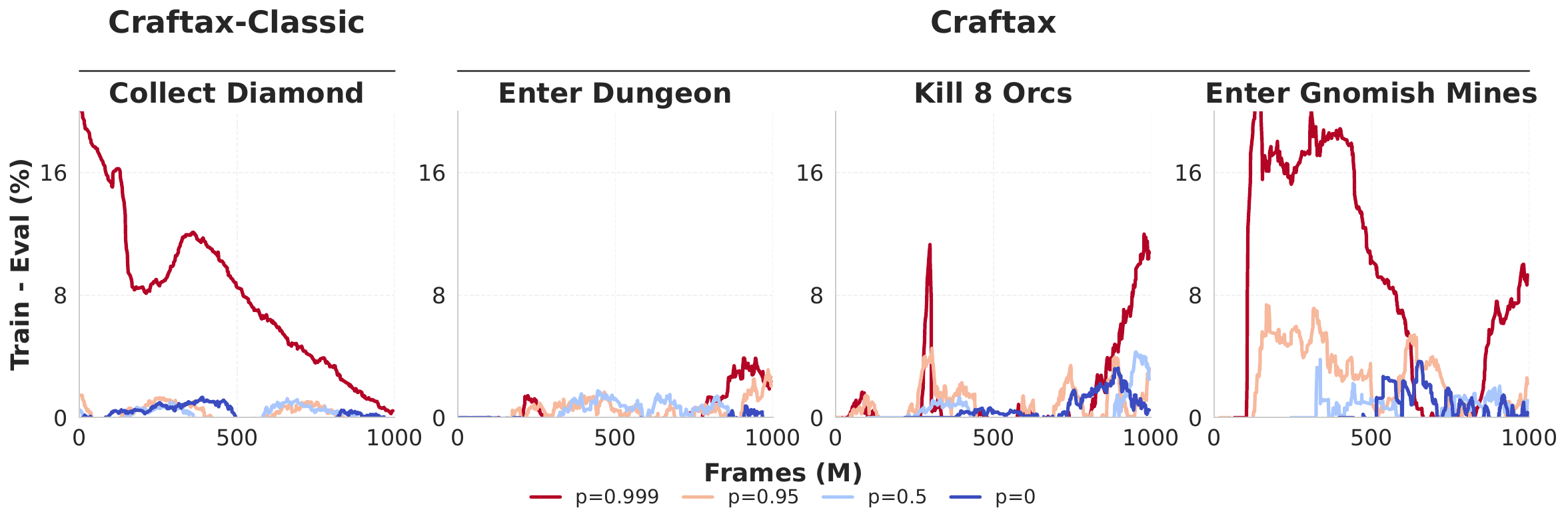}
  \caption{Train-eval gap, measured as training accuracy minus evaluation accuracy, across reset probabilities. At $\alpha = 0.999$ shown in red, a gap of 12--18\% emerges across all tasks; lower reset probabilities maintain near-zero gaps.}
  \label{fig:eval-train-gap}
\end{figure}

\paragraph{Extreme reset causes overfitting to checkpoint configurations.} The relationship between reset probability and performance is not monotonic. While $\alpha = 0$ fails on long-horizon tasks, $\alpha = 0.999$ shown in red often underperforms intermediate values like $\alpha = 0.95$ or $\alpha = 0.99$. Figure~\ref{fig:eval-train-gap} shows the cause: at $\alpha = 0.999$, a train-eval gap of 12--18\% emerges across all tasks, while lower reset probabilities maintain gaps below 2\%. Resetting to the same checkpoint repeatedly means training on identical procedurally generated layouts; the policy overfits to that specific configuration and fails to generalize during evaluation. Intermediate reset values balance sufficient frames at the target skill against diversity from fresh episodes.

\newpage
\subsubsection{Graduation Threshold}
\label{sec:app-graduation-threshold}

\begin{figure}[H]
  \centering
  \includegraphics[width=0.9\textwidth]{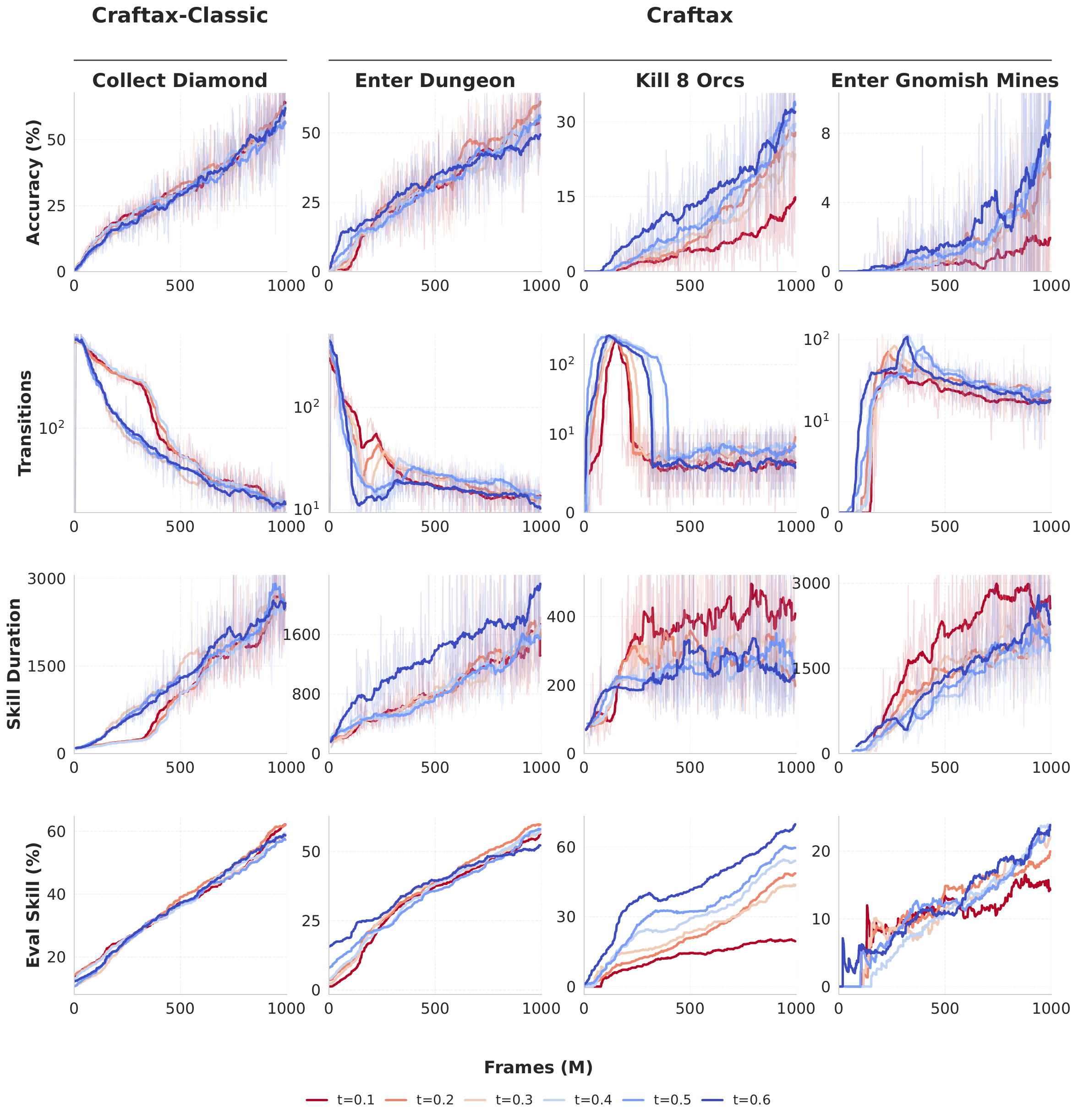}
  \caption{Effect of graduation threshold $\alpha_{\mathrm{grad}}$ across four tasks. Lower thresholds shown in red with $\alpha = 0.1$ advance sooner; higher thresholds shown in blue with $\alpha = 0.5$ require more mastery. \textbf{Row 1:} Overall success. \textbf{Row 2:} Transitions per batch. \textbf{Row 3:} Skill duration. \textbf{Row 4:} Eval skill success.}
  \label{fig:progressive-threshold}
\end{figure}

We ablate the graduation threshold $\alpha_{\mathrm{grad}}$, the success rate at which we advance Frontier Checkpointing to the next skill's preconditions. Figure~\ref{fig:progressive-threshold} shows thresholds from 0.1 to 0.5.

\paragraph{The optimal threshold depends on where the bottleneck lies.} For Collect Diamond, Enter Dungeon, and Enter Gnomish Mines, all thresholds converge to similar final performance because the bottleneck is the target skill itself, not prerequisite quality. Lower thresholds allow reaching the exploration phase sooner, and the additional time spent exploring outweighs any benefit from more polished prerequisites.

Kill 8 Orcs shows the opposite pattern: perfect correlation between threshold and performance. Higher thresholds with $\alpha = 0.5$ shown in blue achieve approximately 30\% accuracy, while lower thresholds with $\alpha = 0.1$ shown in red plateau at approximately 15--20\%. Row 2 reveals the mechanism: higher thresholds produce substantially more transitions early in training. For combat, additional training on prerequisites like entering the dungeon and engaging initial enemies compounds at the target skill through better positioning and health management upon reaching the 8-orc challenge.

\newpage
\subsubsection{Network Sharing}
\label{sec:app-network-sharing}

\begin{figure}[H]
  \centering
  \includegraphics[width=0.9\textwidth]{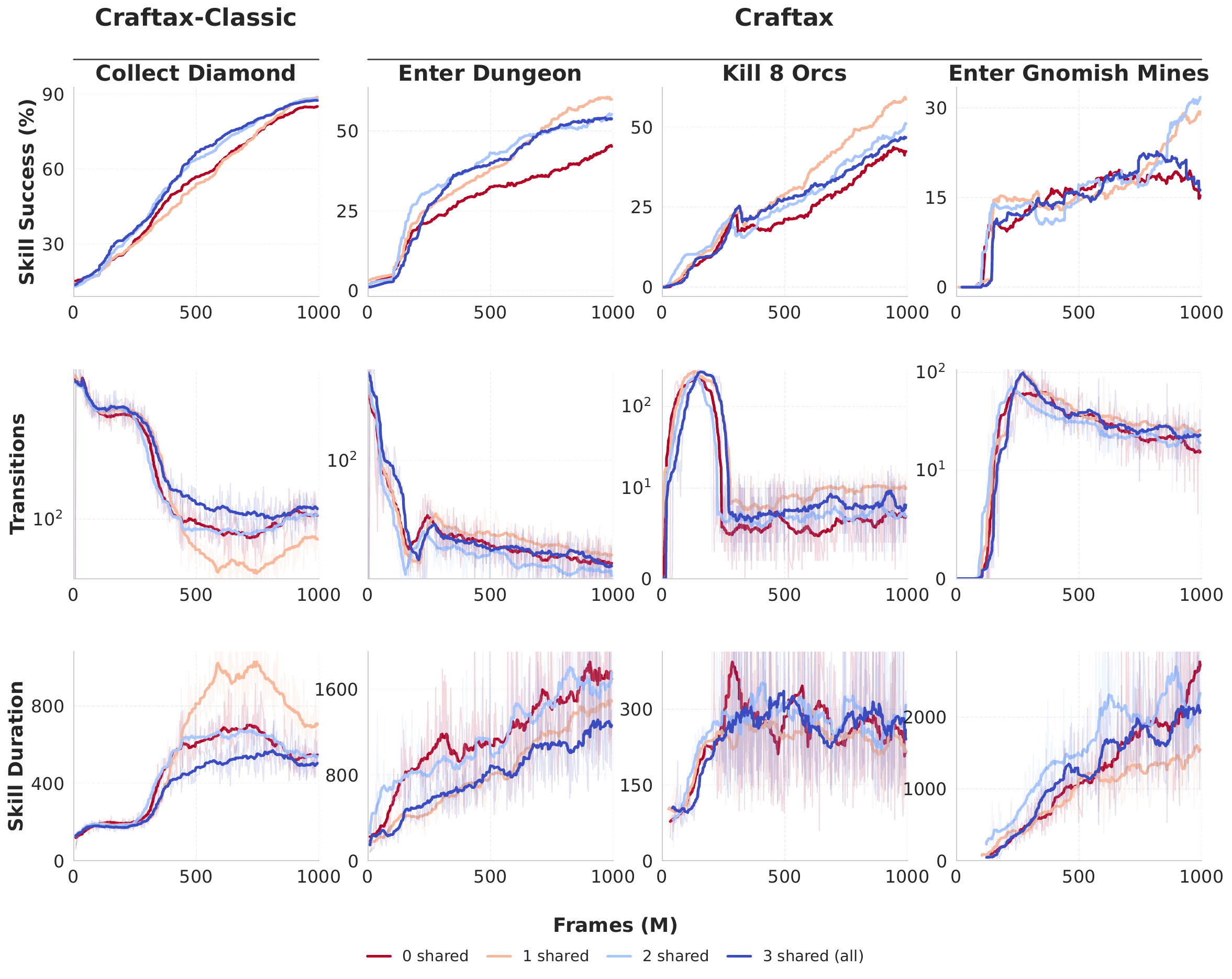}
  \caption{Effect of network sharing across four tasks. Red with 0 shared layers uses separate networks per skill; blue with 3 shared layers uses a single network; intermediate colors share 1--2 layers. \textbf{Row 1:} Skill success. \textbf{Row 2:} Transitions. \textbf{Row 3:} Skill duration.}
  \label{fig:network-sharing}
\end{figure}

We ablate parameter sharing across skills, from fully separate networks with 0 shared layers to a single shared network with 3 shared layers. Figure~\ref{fig:network-sharing} shows the comparison.

\paragraph{Shared early layers transfer features; dedicated final layers prevent interference.} Sharing 1--2 layers outperforms both fully separate and fully shared networks across all tasks. On Kill 8 Orcs, 1--2 shared layers achieve approximately 60\% skill success versus approximately 50\% for both extremes. Shared early layers enable transfer of common features such as spatial reasoning and basic perception, while dedicated final layers allow skill-specific specialization without interference.

Fully separate networks with 0 shared layers shown in red never outperform shared alternatives, even for combat, which requires specialized behaviors distinct from navigation. This suggests low-level features transfer usefully across all skill types, and re-learning them wastes capacity. Enter Dungeon shows the largest benefit from sharing: fully shared networks achieve approximately 60\% skill success versus approximately 50\% for separate networks, as spatial reasoning transfers particularly well across navigation skills.

\newpage
\subsection{Behavioral Analysis}
\label{sec:app-behavioral}

We analyze achievement rates and episode lengths to understand how SCALAR's goal-directed behavior differs from monolithic baselines.

\begin{figure}[H]
  \centering
  \includegraphics[width=0.85\textwidth]{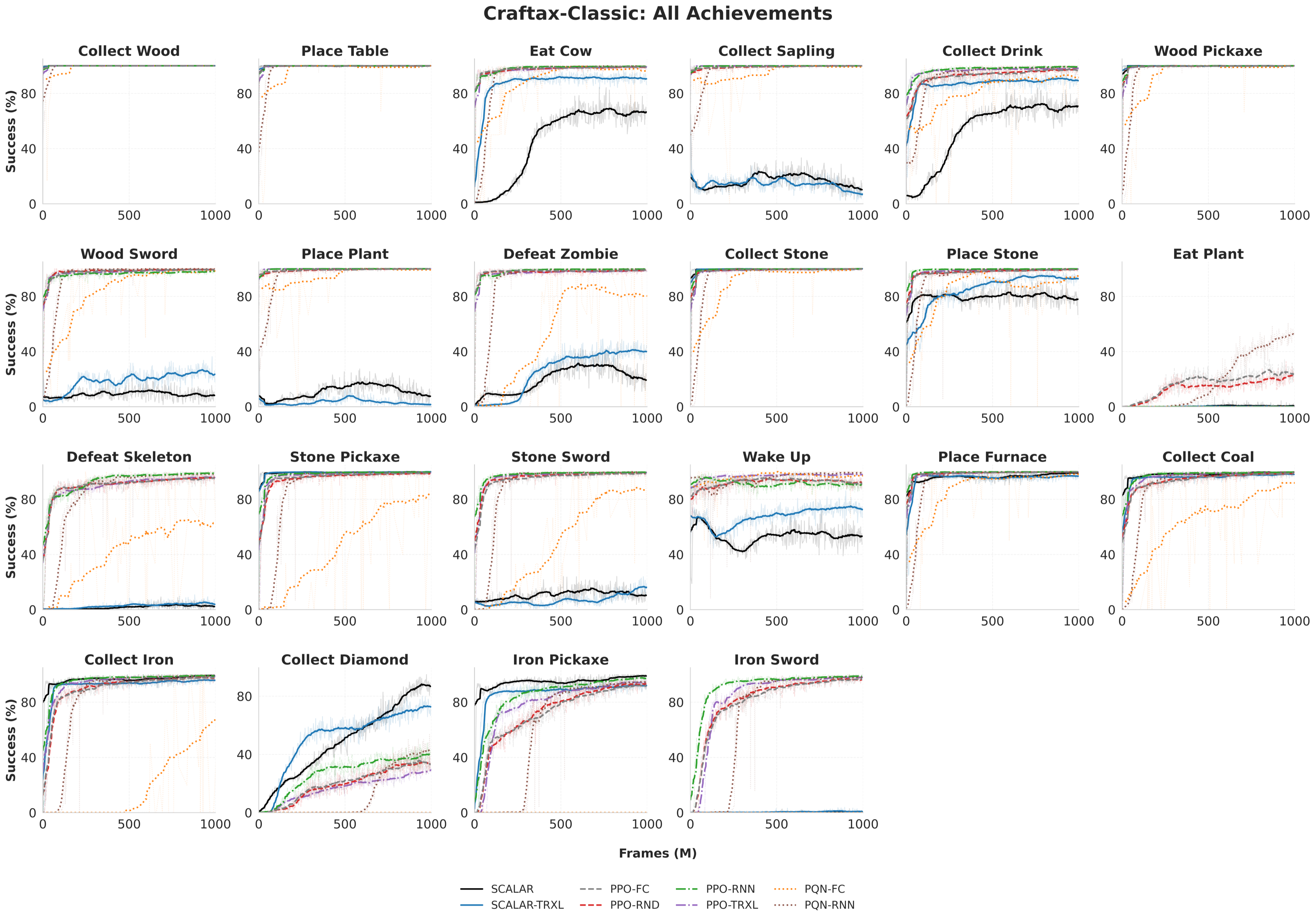}
  \caption{All achievement rates on Craftax-Classic with goal Collect Diamond. SCALAR achieves high rates on task-relevant achievements while avoiding orthogonal ones like sword crafting.}
  \label{fig:all-achievements-classic}
\end{figure}

\begin{figure}[H]
  \centering
  \includegraphics[width=0.85\textwidth]{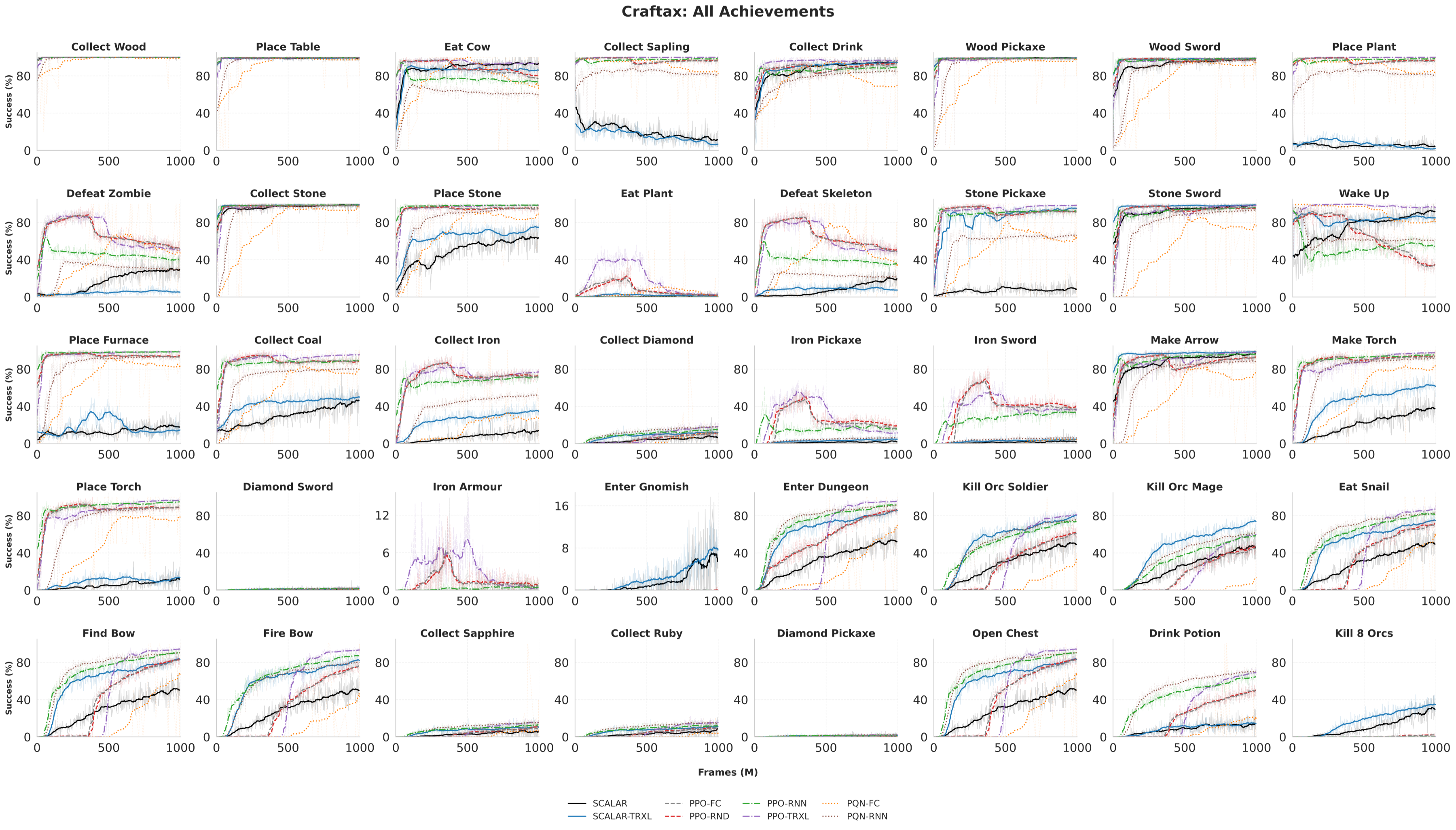}
  \caption{All achievement rates on Craftax with goal Enter Gnomish Mines. SCALAR skips iron tool crafting entirely, using stone tools and arrows instead.}
  \label{fig:all-achievements-craftax}
\end{figure}

\paragraph{SCALAR pursues only task-relevant achievements.} On Craftax-Classic, SCALAR achieves near-zero rates on Make Sword, Make Stone Sword, and Iron Sword, all weapons irrelevant to diamond collection. Baselines achieve 80\%+ on these achievements because their reward signal incentivizes all achievements equally. SCALAR also shows lower rates on Eat Cow and Collect Drink, performing these actions only when necessary for survival rather than proactively. On Craftax, both SCALAR variants skip iron tools entirely, using stone tools and arrows for dungeon combat since stone suffices and iron gathering would delay reaching the target skill. This goal-directed behavior contributes to SCALAR's lower Enter Dungeon rates compared to baselines as shown in Table~\ref{tab:results}: combat experience from fighting zombies and skeletons transfers to dungeon survival, but these auxiliary behaviors are not captured by skill specifications and thus not learned.

\begin{figure}[H]
  \centering
  \includegraphics[width=0.85\textwidth]{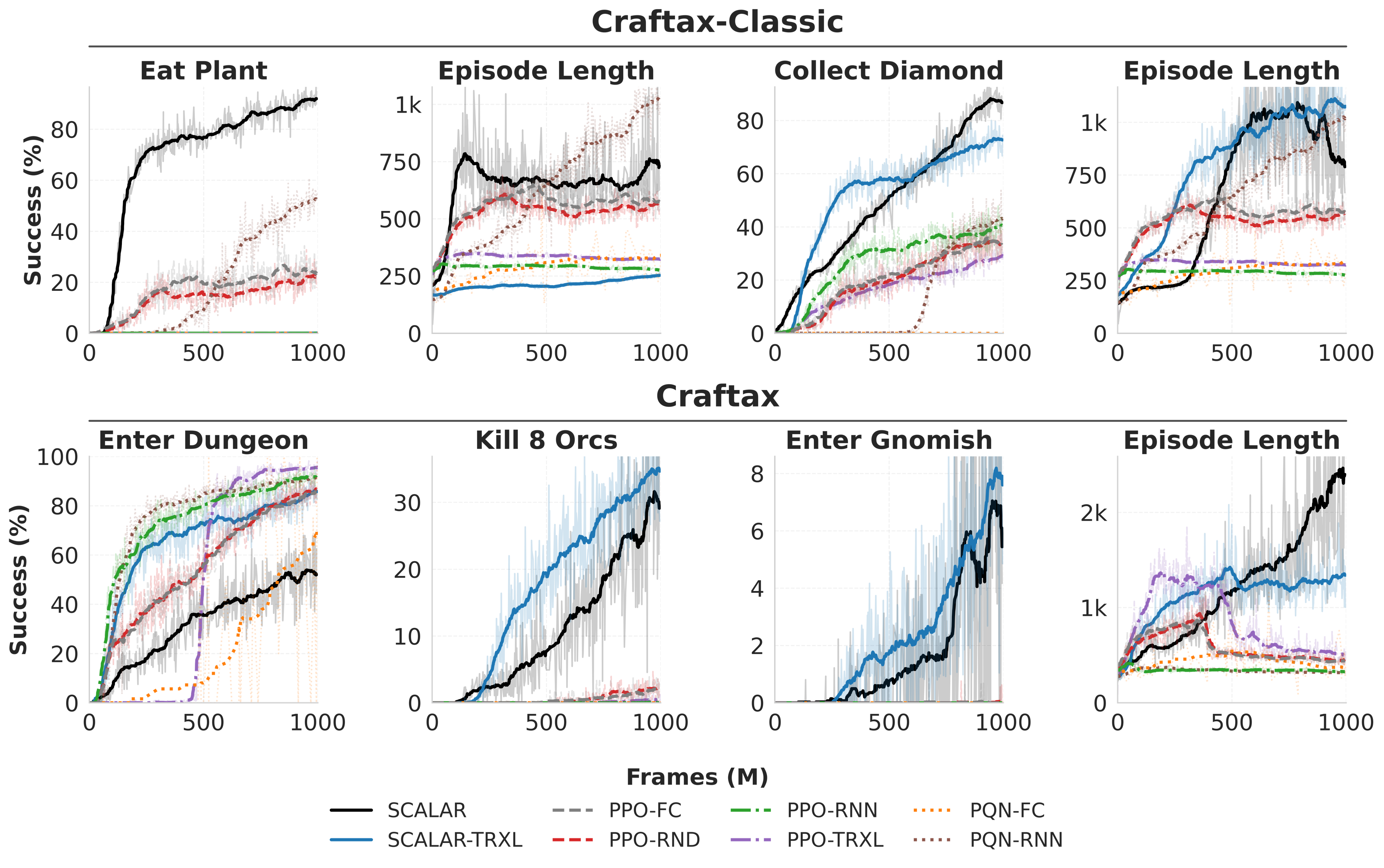}
  \caption{Success rate and episode length across all tasks. Top: Craftax-Classic. Bottom: Craftax. SCALAR's longer episodes reflect task completion; baselines have shorter episodes because they fail early.}
  \label{fig:comprehensive-episode-length}
\end{figure}

\paragraph{Survival is necessary but not sufficient.} On Eat Plant, SCALAR-FC averages approximately 750 frames, learning survival behaviors like sleep, eat, and drink to stay alive during the approximately 600 steps required for plant maturation. Most baselines average approximately 250--500 frames and achieve only approximately 20\% success, dying before the plant matures. PQN-RNN is the exception: it learns comparable survival and reaches similar episode lengths, yet still underperforms SCALAR on Collect Diamond. The difference is focus: PQN-RNN spreads effort across all 21 achievements, while SCALAR's skill structure concentrates training on the target task. On Craftax, the pattern holds: SCALAR episodes extend to approximately 1500--2000 frames for Enter Gnomish Mines, while baselines remain under 1000 frames because they fail before reaching the target, not because they complete faster. SCALAR-TRXL fails on Eat Plant despite identical training procedures, pointing to an unexplained interaction between the transformer architecture and long-wait credit assignment.

\newpage
\subsection{Limitations and Future Work}
\label{sec:app-limitations}

\paragraph{Predefined symbolic space limits representational flexibility.} \ourmethod{} assumes skills can be specified using a fixed symbolic vocabulary, meaning preconditions and postconditions must be expressible within the predefined state encoder and labeling map. This parallels assumptions in classical options frameworks that require hand-designed initiation sets. Three directions could address this limitation. First, learning the symbolic encoder jointly with skill policies, using trajectory analysis to identify which state features are predictive of skill success. Second, expanding the vocabulary online when the LLM proposes skills involving concepts not in the current space. Third, using vision-language models to ground skill specifications directly in observations rather than symbolic states. We note that trajectory analysis partially mitigates this limitation by correcting \emph{quantitative} errors within the symbolic space such as resource counts, even when the \emph{qualitative} structure is fixed.

\paragraph{Fixed skill ordering constrains optimal behavior.} Both FC and TRXL execute skills in a fixed order: the agent explores for the Gnomish Mines entrance only \emph{after} killing 8 orcs. More optimal behavior would interleave exploration with combat, since these subtasks are order-invariant. Future work could combine static planning with an option-critic framework, allowing the agent to select among order-invariant skills dynamically.

\paragraph{Entropy regularization causes spurious actions.} Torches provide no benefit until reaching the Gnomish Mines, yet SCALAR's torch placement rates, while lower than baselines, are not zero. The entropy bonus in PPO prevents any action probability from reaching zero, so suboptimal actions retain probability mass. Action masking based on skill specifications could eliminate these spurious behaviors entirely.

\paragraph{Action masking for safety-critical domains.} If a skill's specification indicates which actions are never required, masking those actions during training would prevent the policy from learning them. Beyond efficiency, this has applications in safety-critical domains: constraining the action space based on symbolic specifications could provide guarantees that certain behaviors cannot occur, even under distributional shift.

\paragraph{SCALAR-TRXL fails on Eat Plant.} Despite identical training procedures, SCALAR-TRXL achieves near-zero success on Eat Plant while SCALAR-FC achieves approximately 90\%. We hypothesize this reflects an interaction between the transformer architecture and the specific credit assignment challenges of long-wait survival tasks, but have not isolated the cause.

\paragraph{Diversity-overfitting tradeoff requires further study.} While the train-eval gap at high reset probabilities shown in Figure~\ref{fig:eval-train-gap} suggests overfitting to checkpoint configurations, we cannot definitively rule out alternative explanations. Future work should investigate whether diversity interventions such as randomizing checkpoints across procedurally generated layouts can recover the benefits of high reset without the overfitting cost.

\paragraph{Frontier Checkpointing requires state serialization.} Frontier Checkpointing assumes the ability to serialize and restore environment state, which is straightforward in simulators but non-trivial elsewhere. For stateful architectures like Transformer-XL, checkpointing also requires saving and restoring the network's memory buffer; under static JAX compilation this adds complexity, so SCALAR-TRXL forgoes checkpointing entirely. Extending checkpointing to settings without native serialization such as real-world robotics would require additional machinery like learned world models or reversible dynamics. Despite this, SCALAR-TRXL achieves 8.2\% on Enter Gnomish Mines without any checkpointing, suggesting memory-based exploration can partially compensate for the frame allocation disadvantage.

\paragraph{Evaluation is limited to domains with reasonable LLM priors.} We evaluate on Craftax and Craftax-Classic, environments where the LLM possesses meaningful prior knowledge about crafting dependencies from Minecraft-like games in training data. While the LLM's priors are imperfect and trajectory analysis corrects systematic overestimates, the qualitative structure of wood leading to tools leading to mining is broadly correct. This choice is deliberate: Craftax provides long-horizon tasks with clear prerequisite chains that stress-test compositional skill learning, unlike standard RL benchmarks that are either short-horizon or solvable by flat policies. However, generalization to domains where LLM priors are systematically \emph{incorrect} rather than merely imprecise remains open. Future work should evaluate on procedurally generated environments that violate common-sense assumptions, testing whether trajectory analysis can correct qualitative errors in addition to quantitative ones.

\newpage
\section{Extended Related Work}
\label{app:extra_related_work}

This appendix provides additional context on related work, extending the discussion in the main text.

\subsection{Unsupervised Skill Discovery}

Long-horizon tasks benefit from learning a collection of skills rather than a single monolithic policy. Traditional skill discovery methods \citep{bagaria2021skill,ecoffet2019go} identify useful behavioral primitives through exploration and graph-based representations. \citep{evans2023creating} develops multi-level skill hierarchies for navigation in maze-like domains, while substantial research frames skill emergence as maximizing dependence between states and skill labels via mutual information \citep{vic,diayn,dads,hansen2019fast,liu2021aps,cic}. However, discriminator-based mutual information objectives can saturate once a classifier perfectly separates skills, often yielding behaviors that differ only in subtle, non-salient ways \citep{diayn}. To promote more behaviorally distinct skills, recent methods replace mutual information with Wasserstein dependency measures \citep{ozair2019wasserstein,he2022wasserstein,lsd,csd,metra}, pairing the objective with task-relevant metrics such as Euclidean distance in state space \citep{lsd} or controllability-aware distances that favor rare transitions \citep{csd}. Disentangled representations can further improve skill separation by isolating independent factors of variation in the state space~\citep{zabounidis2026disentangled}. These methods discover skills without external guidance; \ourmethod{} instead leverages LLM priors to propose task-relevant skill structures directly.

\subsection{Classical Hierarchical RL}
\label{app:classical_hrl}

Rather than discovering skills from scratch, classical hierarchical RL provides structure through hand-designed abstractions. The options framework \citep{SUTTON1999181} formalized temporally extended actions with initiation sets, policies, and termination conditions. Option-Critic \citep{bacon2016optioncriticarchitecture} enabled end-to-end option learning via policy gradient, eliminating the need for hand-designed option structures. HIRO \citep{nachum2018hiro} introduced goal-conditioned hierarchies with off-policy correction for improved sample efficiency. Feudal Networks \citep{vezhnevets2017feudal} learn hierarchical policies where a manager sets goals for workers. SDRL \citep{sdrl2019} integrates symbolic planning with options-based RL, using a planner-controller-meta-controller architecture where symbolic knowledge guides subtask scheduling. \ourmethod{} shares the options framework's use of explicit preconditions and termination conditions, but sources these specifications from LLM priors rather than hand-designed symbolic knowledge, then refines them through trajectory analysis.

\subsection{LLM-Generated Reward Functions}

Reinforcement learning for long-horizon tasks faces significant challenges in defining precise reward functions that effectively guide learning without introducing unintended behaviors \cite{hadfield2017inverse, krakovna2020avoidingeffectsconsideringfuture, everitt2021rewardtamperingproblemssolutions}. Intrinsic motivation and reward shaping techniques provide additional unsupervised learning signals \citep{pathak2017curiosity}, but recent research has investigated leveraging LLMs to automatically construct reward functions from task descriptions. Early explorations \citep{read_and_reward,xie2023text2reward,ma2023eureka} focused on generating reward functions for complete tasks without direct involvement in agent interactions. \citep{deng2025reward} extends this approach by enabling LLMs to provide dynamic reward adjustments based on agent interactions, assigning positive feedback to beneficial actions and negative feedback to detrimental ones. \citep{liu2024worldmodelshintslarge} integrates LLM-generated subgoal hints into model rollouts, providing intrinsic rewards for goal completion. These methods use LLMs for reward specification; \ourmethod{} goes further by having LLMs specify full skill structures including preconditions, postconditions, and compositional dependencies.

\subsection{LLM-Guided Skill Specification}

Beyond reward functions, LLMs can specify complete skill structures. Methods in this space transform high-level goals into skill definitions with dense rewards and termination conditions \citep{leagueorginal,li2024league,mao2025skill}, or generate subgoal sequences before training \citep{shek2025optiondiscoveryusingllmguided}. Motif \citep{klissarov2024motif} trains neural reward models from LLM annotations of trajectory pairs, enabling RL agents to learn behaviors described in natural language. MaestroMotif \citep{klissarov2025maestromotif} extends this to hierarchical skill learning, where humans provide natural language skill descriptions, Motif trains reward models for each skill, and an LLM-generated Python policy selects which skill to execute. These approaches typically follow a one-shot, feedforward plan that is not refined through interaction. Unlike MaestroMotif, \ourmethod{} has the LLM propose skill decompositions rather than requiring human-authored descriptions, uses LLM-generated reward code directly rather than learned reward models, and refines specifications through trajectory analysis.

\subsection{Symbolic Planning and RL}

The skill structures we generate connect to classical AI planning. Symbolic operators with preconditions and effects enable STRIPS-style planning toward goals \citep{fikes1971strips}. Prior work integrates planning with hierarchical RL: SDRL \citep{sdrl2019} uses symbolic knowledge to guide subtask scheduling, PEORL \citep{peorl2018} uses hand-authored action models to guide option execution, and taskable RL \citep{illanes2020symbolic} uses symbolic plans as instructional scaffolding. These methods assume fixed, high-fidelity symbolic models. More recently, SayCan \citep{saycan} uses LLMs to propose high-level plans grounded through learned affordance models, and League++ \citep{li2024league} has an LLM propose plans verified by an A* planner. \ourmethod{} combines these threads: we use LLMs to specify STRIPS-like operators that classical planners sequence, but unlike prior work, we refine these specifications through trajectory analysis during training rather than assuming they are correct from the start.

\subsection{Refining LLM Priors through Interaction}

LLM-based skill specification methods rely on static, one-shot planning that cannot adapt when initial LLM context proves insufficient. Environment interaction becomes essential when LLM understanding is incomplete or incorrect. LLM self-improvement through environment feedback succeeds in coding \citep{madaan23SelfRefine,jin2025reveal} and planning agents \citep{wang2023voyager,wu2024agentkit} using generate-evaluate-reflect cycles. \citep{wu2024agentkit} shows LLMs gather environment dynamics from low-level interactions, and Skill Set Optimization \citep{nottingham2024skill} extracts text-based skill descriptions from successful trajectories and refines them through interaction, though both remain within the LLM-as-agent paradigm where skill execution is handled by the LLM itself. \ourmethod{} refines skill specifications through trajectory analysis while using RL-trained policies for execution, combining the knowledge encoded in language models with the grounding provided by environmental interaction.

\newpage
\FloatBarrier
\section{Experimental Details}
\label{app:exp_setup}\label{app:implementation}

\subsection{Observation Space}
\label{sec:app-encoder}

Craftax-Classic provides symbolic observations: an egocentric $7{\times}9$ local map containing blocks and mobs, nearest-block offsets, inventory, and vitals. We use this parsed state directly as our encoder, treating the current observation as sufficient for the symbol, so that $z_t = \Phi(o_t)$ and $L(z_t)\subseteq \mathcal{F}$.

We augment the symbolic observation with relative position offsets to key entities including nearest blocks and mobs to reduce aliasing from local views and to match the input space used for all baselines in Section~\ref{sec:exp_setup}. These offsets encode the displacement from the agent to the nearest instance of each relevant block or mob type in a fixed-radius neighborhood.

\subsection{SCALAR Implementation}

\paragraph{Ephemeral Skill Handling.}
\label{sec:app-ephemeral}

A skill $\sigma$ is \emph{ephemeral} when its gains do not persist across time or position, such as $\textsc{Near(CraftingTable)}$ after walking away. The LLM decides ephemerality from symbolic rollouts. During planning, ephemeral skill prerequisites are substituted by their \emph{requirements}: if \textsc{PlaceCraftingTable} requires 4 wood, then \textsc{CraftPickaxe} requires $\{6\text{ wood}, 1\text{ iron}\}$ instead of $\{2\text{ wood}, 1\text{ iron}, \textsc{Near(CraftingTable)}\}$. This ensures the frontier reflects persistent capabilities rather than transient intermediates.

\paragraph{Skill Composition.}
\label{sec:app-jax}

Given a library $\Sigma$, we compose skills to reach frontier states that satisfy a proposed skill's requirements. We approximate the reachable set by tracing backwards from the proposed skill's requirements through the dependency graph, using breadth-first traversal to determine an execution order where all prerequisites appear in earlier layers. This produces a tractable subset of reachable facts sufficient for feasibility and novelty checks.

\paragraph{Architecture.}
\label{sec:app-moe}

SCALAR uses the same architectures as baselines but with a mixture-of-experts structure: skills share a common backbone while maintaining separate actor-critic heads, enabling independent learning without interference or catastrophic forgetting. For SCALAR-FC, which uses a 4-layer MLP with 512 hidden units, the first layer is shared with 3 skill-specific layers per expert. For SCALAR-TRXL, which uses 4 transformer layers with 256 hidden units and 4 heads, the entire transformer backbone is shared with 2 skill-specific linear layers per expert. SCALAR-FC uses Frontier Checkpointing; SCALAR-TRXL does not. See Appendix~\ref{sec:app-compute} for details on this architectural constraint and Appendix~\ref{sec:app-architecture} for performance comparisons.

\paragraph{Training Protocol.}

Skills train with PPO until reaching 30\% success rate, then the curriculum advances. Goal tasks including diamond collection and Gnomish Mines entry train to 99\% to consume the remaining training budget. Frontier Checkpointing resets to saved frontier states with probability $p=0.9$ during training, ensuring most frames target the current skill rather than re-executing prerequisites. Trajectory analysis triggers upon first nonzero success, analyzing $K=10$ successful trajectories to refine operator specifications. If a skill fails to reach nonzero success within its training budget, trajectory analysis triggers on failure trajectories to identify missing prerequisites. Each skill receives a budget of $10^8$ frames, one-tenth of the total training budget.

\paragraph{LLM Integration.}

We use GPT-5 Thinking Medium for skill proposal and trajectory analysis, with temperature $T=0$ for deterministic outputs. Each query includes environment specifications, the current knowledge base, and descriptions of existing skills. Generated reward and completion functions undergo syntax verification, JAX compatibility checking, and logical consistency tests before deployment.

\subsection{Network Architectures}
\label{sec:app-network-architectures}

All PPO baseline and SCALAR architectures follow the implementations from \citet{matthews2024craftax}, available at \url{https://github.com/MichaelTMatthews/Craftax_Baselines}. PQN baselines follow the implementation from \citet{gallici2025simplifyingdeeptemporaldifference}, available at \url{https://github.com/mttga/purejaxql}. The Transformer-XL implementation is based on \url{https://github.com/Reytuag/transformerXL_PPO_JAX}. We describe each architecture below.

\subsubsection{PPO-FC (Feedforward)}
\label{sec:app-ppo-fc}

The feedforward baseline uses a 4-layer MLP with separate actor and critic networks. Both networks use three hidden layers of 512 units with Tanh activation, followed by an output layer. The actor outputs $|\mathcal{A}|$ units with categorical distribution; the critic outputs a single value. All layers use orthogonal initialization with gain $\sqrt{2}$ except output layers. For the actor output layer we use gain 0.01; for the critic output layer we use gain 1.0. Total parameters: approximately 2.1M.

\subsubsection{PPO-RND (Feedforward + Random Network Distillation)}
\label{sec:app-ppo-rnd}

PPO-RND extends PPO-FC with Random Network Distillation exploration bonuses. RND uses a fixed randomly initialized \textbf{target network} and a trainable \textbf{predictor network} that attempts to predict the target network's output. The exploration bonus equals the prediction error measured as MSE between predictor and target outputs, which is high for novel states. Both networks use three 256-unit ReLU layers with 32-dimensional outputs. The intrinsic reward is scaled by coefficient 1.0.

\subsubsection{PQN-FC (Feedforward Q-Network)}
\label{sec:app-pqn-fc}

PQN uses value-based learning without actor-critic separation. The Q-network is a 4-layer MLP with 1024 hidden units, BatchRenorm after each layer, and ReLU activation, outputting Q-values for each action. BatchRenorm provides stable normalization critical for value-based methods with high variance in target values. Policy is derived via $\epsilon$-greedy selection.

\subsubsection{PQN-RNN (Recurrent Q-Network)}
\label{sec:app-pqn-rnn}

PQN-RNN extends PQN-FC with LSTM memory. The encoder uses a single 512-unit layer with LayerNorm and ReLU, followed by an OptimizedLSTMCell with 512 hidden units. LayerNorm replaces BatchRenorm for stable gradients through time. Hidden state resets at episode boundaries. The LSTM output feeds a linear layer producing Q-values.

\subsubsection{PPO-RNN (Recurrent)}
\label{sec:app-ppo-rnn}

PPO-RNN uses a GRU cell with 512 hidden units preceded by a 512-unit ReLU embedding layer. Actor and critic heads each use two 512-unit ReLU layers before their output layers. Hidden state resets at episode boundaries; training uses 64-timestep sequences with truncated backpropagation through time.

\subsubsection{PPO-TRXL (Transformer-XL)}
\label{sec:app-ppo-trxl}

The Transformer-XL baseline uses self-attention with relative positional embeddings and memory caching. The backbone consists of a 256-dim encoder projection followed by 2 transformer layers. Each layer uses pre-norm, multi-head relative positional attention with 8 heads and 256 dim, gated residual connections with bias 2.0, and a 256-dim FFN with GELU. Memory caches 128 key-value pairs per layer. Actor and critic heads each use two 256-unit Tanh layers.

\subsubsection{SCALAR Architectures}
\label{sec:app-scalar-arch}

SCALAR uses mixture-of-experts variants with shared backbones and per-skill heads. \textbf{SCALAR-FC} shares the first dense layer of 512 units across skills, with 3 skill-specific layers per expert plus independent actor-critic heads. \textbf{SCALAR-TRXL} shares the entire transformer backbone, with 2-layer skill-specific actor and critic heads. During training, only current skill head parameters update while the backbone receives gradients from all skills, preventing catastrophic forgetting.

\paragraph{Model Capacity.}
PPO-FC uses approximately 2.1M parameters. SCALAR-FC uses approximately 2.1M shared backbone parameters plus approximately 0.8M per skill for actor-critic heads. With 8--12 skills for the diamond task, total capacity ranges from approximately 8.5M to 11.7M parameters, though only approximately 3.0M are active during any single skill's forward pass. SCALAR-TRXL shares the approximately 1.2M parameter transformer backbone with approximately 0.3M per-skill heads. Total capacity is comparable to or smaller than monolithic TRXL baselines when accounting for skill reuse across tasks.

\subsection{Training Hyperparameters}
\label{sec:app-hyperparams}

Table~\ref{tab:ppo_hyperparams} shows hyperparameters for PPO-based methods. PPO-TRXL uses a higher discount factor, lower entropy coefficient, and longer rollouts to accommodate transformer memory.

\begin{table}[h]
\centering
\small
\begin{tabular}{@{}lcc@{}}
\toprule
\textbf{Hyperparameter} & \textbf{PPO-FC/RNN/RND} & \textbf{PPO-TRXL} \\
\midrule
Learning rate & $2 \times 10^{-4}$ & $2 \times 10^{-4}$ \\
Discount factor ($\gamma$) & 0.99 & 0.999 \\
GAE lambda ($\lambda$) & 0.8 & 0.8 \\
Clip coefficient ($\epsilon$) & 0.2 & 0.2 \\
Entropy coefficient & 0.01 & 0.002 \\
Value function coefficient & 0.5 & 0.5 \\
Maximum gradient norm & 1.0 & 1.0 \\
Update epochs & 4 & 4 \\
Minibatches & 8 & 8 \\
Steps per rollout & 64 & 128 \\
Parallel environments & 1024 & 1024 \\
Total timesteps & $10^9$ & $10^9$ \\
\bottomrule
\end{tabular}
\caption{PPO hyperparameters. All methods use linear learning rate annealing and optimistic resets with ratio 16.}
\label{tab:ppo_hyperparams}
\end{table}

PPO-RND uses predictor network learning rate $3 \times 10^{-4}$, reward coefficient 1.0, and embedding dimension 32.

Table~\ref{tab:pqn_hyperparams} shows hyperparameters for PQN methods. PQN-FC uses single-step updates while PQN-RNN uses multi-step rollouts with TD-lambda.

\begin{table}[h]
\centering
\small
\begin{tabular}{@{}lcc@{}}
\toprule
\textbf{Hyperparameter} & \textbf{PQN-FC} & \textbf{PQN-RNN} \\
\midrule
Learning rate & $1 \times 10^{-4}$ & $3 \times 10^{-4}$ \\
Discount factor ($\gamma$) & 0.99 & 0.99 \\
TD-lambda ($\lambda$) & 0.0 & 0.5 \\
$\epsilon$-greedy (start $\rightarrow$ end) & $0.1 \rightarrow 0.005$ & $1.0 \rightarrow 0.005$ \\
Maximum gradient norm & 1.0 & 0.5 \\
Update epochs & 1 & 4 \\
Minibatches & 1 & 4 \\
Steps per rollout & 1 & 128 \\
\bottomrule
\end{tabular}
\caption{PQN hyperparameters. Both use 1024 parallel environments, $10^9$ total timesteps, and linear LR annealing.}
\label{tab:pqn_hyperparams}
\end{table}

SCALAR uses the same base hyperparameters as the corresponding PPO baseline, with skill success threshold 80\% for standard skills and 99\% for goal tasks, and frontier checkpoint probability 0.9.

\subsection{Baseline Training}
\label{sec:app-baselines}

All baselines receive reward for all 22 Craftax achievements equally, following the standard benchmark setup. This spreads learning signal across tasks like sword crafting that are irrelevant to the target objective.

\newpage
\subsection{Computational Details}
\label{sec:app-compute}

Training uses 1024 parallel environment instances with vectorized JAX execution. JIT compilation eliminates Python interpreter overhead; persistent compilation caching reduces startup time across runs. Each 1B-frame experiment completes in approximately 1 hour on a single H100 GPU. All experiments use 5 random seeds.

\paragraph{LLM Cost.}

A key advantage of SCALAR is that the LLM is not invoked during RL training. LLM calls occur only between training runs for skill proposal, reward code generation, and trajectory analysis. This amortizes inference cost over millions of environment steps. For all experiments reported, total LLM cost was \$2.83 with 1.429M input tokens and 125k output tokens.

\paragraph{JAX Compilation and Frontier Checkpointing Constraints.}

End-to-end JAX RL requires static computation graphs: all tensor shapes and control flow must be known at compile time. This enables aggressive optimization but constrains certain operations.

Frontier Checkpointing saves environment state to resume training from frontier states. This works straightforwardly for feedforward policies: we store a snapshot of environment state and restore it when needed. The network has no hidden state, so restoration is complete.

For transformer-based policies like TRXL, checkpointing is more complex. The transformer maintains a memory buffer of past key-value pairs that conditions its predictions. Restoring to a saved environment state also requires restoring the memory buffer to its corresponding value; otherwise the network conditions on a mismatched context. Under static JAX compilation, this memory is part of the compiled computation graph. Dynamically injecting saved memory states would require either recompilation per checkpoint or a more complex memory management scheme that maintains consistency between environment state and network memory.

Rather than introduce this complexity, SCALAR-TRXL forgoes Frontier Checkpointing. This means TRXL must re-execute prerequisites from episode start, spending more frames on prerequisite skills than SCALAR-FC. Despite this frame allocation disadvantage, TRXL achieves competitive performance on deep prerequisite chains as shown in Table~\ref{tab:results}.

\subsection{Environment Selection}
\label{sec:app-env-selection}

Long-horizon RL benchmarks with deep compositional structure are rare. SCALAR requires environments where tasks decompose into prerequisite chains with symbolic structure LLMs can reason about, sparse rewards create genuine exploration challenges, and computational tractability permits extensive experimentation.

\textbf{Minecraft} is conceptually ideal. It offers deep crafting trees, open-ended exploration, and rich symbolic structure. However, it is computationally prohibitive: training 1B frames in Minecraft would require weeks of wall-clock time versus approximately 1 hour in Craftax. Minecraft experiments remain valuable future work, but would limit the ablation studies and seed counts we report here.

\textbf{NetHack} offers long-horizon challenges but less compositional crafting structure. MaestroMotif~\citep{klissarov2025maestromotif} addresses NetHack with only 5 skills in a simple oscillating pattern: descend, explore, ascend, repeat. The skill graph is shallow; skills do not have resource dependencies or crafting prerequisites. This makes NetHack less suitable for evaluating compositional skill learning where each skill's feasibility depends on completing prior skills.

\textbf{Craftax} combines Minecraft-style crafting with tractable computation. Diamond collection requires a 10+ step prerequisite chain: gather wood, craft table, craft pickaxe, mine stone, craft furnace, smelt iron, craft iron pickaxe, find and mine diamond. Each step consumes resources and produces new capabilities, creating genuine compositional dependencies. The full Craftax environment adds combat requirements such as killing 8 orcs to enter the Gnomish Mines, and 9 procedurally generated floors. JAX implementation enables 1B frames in approximately 1 hour, permitting thorough ablation studies.

\subsection{Baseline Selection}
\label{sec:app-baseline-selection}

\paragraph{Monolithic RL Baselines.}

We compare against PPO and PQN variants trained on all achievements simultaneously. This is the standard benchmark setup~\citep{matthews2024craftax}. These baselines show what RL achieves without LLM-guided decomposition. SCALAR benefits from goal-directed reward focus by training only skills relevant to the target, but this structural advantage is part of our contribution: we show that LLM-guided decomposition provides this benefit automatically.

Our shared-network ablation in Section~\ref{sec:main_results} provides an intermediate point: when all layers are shared, SCALAR's mixture-of-experts architecture collapses to a standard PPO-FC network, but still uses the LLM-generated execution graph for reward shaping. This becomes a highly structured curriculum, giving reward for one task at a time rather than all achievements simultaneously. This ablation still outperforms standard PPO as shown in Table~\ref{tab:results}, suggesting that structured reward sequencing, even without modular networks, provides value. The full SCALAR method with partial layer sharing performs better still, indicating that both components contribute.

\paragraph{Hierarchical RL Baselines.}

We do not compare against traditional hierarchical RL methods such as Option-Critic, HIRO, or Feudal Networks for two reasons. First, computational tractability: these methods lack JAX implementations and cannot achieve the throughput required for 1B-frame experiments. Similar challenges are reported by \citet{sol2026}, who note that adapting hierarchical baselines to high-throughput settings would require months of wall-clock time per experiment. Second, methodological mismatch: these methods discover skills through exploration or learn option policies end-to-end, whereas SCALAR uses LLM-specified skill structures with RL-trained policies. A fair comparison would require adapting these methods to use the same symbolic skill definitions, which would conflate implementation quality with method quality. We implemented all methods with mature JAX codebases that could scale to our experimental regime.

\newpage
\FloatBarrier
\section{Prompt Details}
\label{sec:app-prompt-details}

SCALAR's knowledge base generation uses a multi-phase LLM pipeline to transform a game tutorial into structured skill definitions with executable code. Each phase produces intermediate JSON that feeds into the next. We present the complete prompts for Craftax-Classic, followed by the floor analysis adaptation for full Craftax.

\subsection{Phase 1: Skill Extraction}
\label{sec:app-prompt-phase1}

The first phase converts a game tutorial into structured skill definitions. The LLM receives the tutorial text along with the specification of the symbolic state space.

\begin{lstlisting}[basicstyle=\ttfamily\scriptsize]
Convert the following Craftax Classic game tutorial into a structured
JSON knowledge base. The knowledge base will be used by an RL agent
to decide which skills to learn next.

# Symbolic State Space

The agent tracks progress through these state types:

**Inventory items** (type: "inventory"):
- wood, stone, coal, iron, diamond, sapling, arrows, torches
- wood_pickaxe, stone_pickaxe, iron_pickaxe (0 or 1)
- wood_sword, stone_sword, iron_sword (0 or 1)

**Achievements** (type: "achievement", prefix: "achievement:"):
COLLECT_WOOD, PLACE_TABLE, EAT_COW, COLLECT_SAPLING, COLLECT_DRINK,
MAKE_WOOD_PICKAXE, MAKE_WOOD_SWORD, PLACE_PLANT, DEFEAT_ZOMBIE,
COLLECT_STONE, PLACE_STONE, EAT_PLANT, DEFEAT_SKELETON,
MAKE_STONE_PICKAXE, MAKE_STONE_SWORD, WAKE_UP, PLACE_FURNACE,
COLLECT_COAL, COLLECT_IRON, COLLECT_DIAMOND, MAKE_IRON_PICKAXE,
MAKE_IRON_SWORD

# Gain Schema

Each skill has gains expressed as:
```json
"gain_key": {
  "type": "inventory|achievement|intrinsic",
  "expression": "lambda n: ...",
  "description": "optional"
}
```

Expression patterns:
- **Count-based** (wood, stone, arrows): `"lambda n: n"` (gain n items)
- **Binary** (wood_pickaxe, stone_sword, etc.): `"lambda n: 1"`
- **Achievements**: `"lambda n: 1"` (binary)

# Tutorial Information

```markdown
{tutorial_text}
```

# Task

Generate a JSON knowledge base with:

1. **tutorial_context**: Preserve the tutorial information.

2. **skills**: List of all skills for Craftax Classic.

   Each skill has ONE primary goal. Do not combine multiple distinct
   goals into a single skill.

   For each skill, provide:
   - skill_name: brief descriptive name
   - description: what the skill does and why it's useful
   - gain: what inventory/achievements the skill provides
     - Use proper gain keys (inventory items, "achievement:X")
     - expression: "lambda n: n" for counts, "lambda n: 1" for binary

   Include skills for:
   - Resource gathering (wood, stone, coal, iron, diamond, sapling)
   - Tool crafting (wood/stone/iron pickaxes and swords)
   - Survival (placing table, furnace, eating, drinking, sleeping)
   - Combat (defeating zombies, skeletons)
   - Plant mechanics (place plant, eat plant)

Do NOT include a `prerequisites` field; dependencies are inferred
later from requirements/consumption.

Return ONLY valid JSON.
\end{lstlisting}

\newpage
\subsection{Phase 2: Prerequisite and Consumption Inference}
\label{sec:app-prompt-phase2}

The second phase determines what each skill requires to execute and what resources it consumes. The LLM receives the skill list from Phase 1 along with the tutorial text.

\begin{lstlisting}[basicstyle=\ttfamily\scriptsize]
You are analyzing prerequisites for a skill in Craftax Classic.

# All Available Skills
```
{all_skills_json}
```

# Current Skill to Analyze
```json
{current_skill_json}
```

# Tutorial Context
```
{tutorial_text}
```

# Task

## Step 0: Understand the Skill
Read the skill name and description carefully.
- A Skill is a reinforcement learning policy executed over
  potentially a long horizon
- Skill requirements are the recommended prerequisites enabling
  successful execution of the skill, and are fulfilled by other
  skills prior to executing this skill

## Step 1: Extract Relevant Tutorial Information
Read the tutorial context and extract ALL information relevant to
achieving the skill:
- What does the tutorial say about how to achieve this skill?
- What resources does the tutorial say are consumed?

## Step 2: Determine Requirements and Consumption
Based on the tutorial information:

**Requirements**: Everything needed for safe and successful execution
of this skill.
- Include both necessary and recommended requirements.
- Consider what requirements are intuitively useful to have even if
  not strictly necessary.
- Let X be some requirement Y be the state after executing the skill.
  We should only include X if it is useful to go from the skill's
  initial state to Y. Whether it is only useful after reaching Y is
  irrelevant.
- Do not factor in the usefulness of a requirement for future skills
  after reaching Y.
- Do NOT include transitive dependencies.

**Consumption**: Only what gets used up during execution.
- Include any of the beforelisted requirements that are likely to be
  consumed during the execution of the skill.
- For variable consumption, use a realistic estimate, leaning
  slightly higher for a safe margin.
- Do NOT include durable equipment that remains after use.

Requirements are a SUPERSET of consumption.

## Step 3: Determine Ephemeral Status
Is the primary gain a PLACEMENT or PROXIMITY effect in the game world?
- Ephemeral = true: Spatial effect (places object, creates adjacency).
- Ephemeral = false: Inventory/Stat change.

# Format

After completing your analysis above, return the JSON:

```json
{
  "skill_name": "",           # Keep same
  "description": "",          # Keep same
  "requirements": {},         # gain_key -> "lambda n: a*n + b"
  "consumption": {},          # gain_key -> "lambda n: a*n + b"
  "gain": {},                 # Keep same
  "ephemeral": false          # true if spatial effect
}
```

Lambda format: `"lambda n: a*n + b"` where n=executions.
Each key must match a gain key from available skills.

Important:
- Tools in Craftax Classic are binary items (wood_pickaxe,
  stone_pickaxe, iron_pickaxe, wood_sword, stone_sword, iron_sword)
  - they are either present (1) or not (0).
\end{lstlisting}

\newpage
\subsection{Phase 3: Reward Function and Code Generation}
\label{sec:app-prompt-phase3}

The third phase translates skill specifications into executable JAX code in two steps.

\paragraph{Step 3a: Dense Reward Design.}

\begin{lstlisting}[basicstyle=\ttfamily\scriptsize]
# All factors

Environment definitions for Craftax Classic:
```python
class BlockType(Enum):
    INVALID = 0
    OUT_OF_BOUNDS = 1
    GRASS = 2
    WATER = 3
    STONE = 4
    TREE = 5
    WOOD = 6
    PATH = 7
    COAL = 8
    IRON = 9
    DIAMOND = 10
    CRAFTING_TABLE = 11
    FURNACE = 12
    SAND = 13
    LAVA = 14
    PLANT = 15
    RIPE_PLANT = 16

@struct.dataclass
class Inventory:
    wood: int = 0
    stone: int = 0
    coal: int = 0
    iron: int = 0
    diamond: int = 0
    sapling: int = 0
    wood_pickaxe: int = 0  # Binary: 0 or 1
    stone_pickaxe: int = 0  # Binary: 0 or 1
    iron_pickaxe: int = 0  # Binary: 0 or 1
    wood_sword: int = 0  # Binary: 0 or 1
    stone_sword: int = 0  # Binary: 0 or 1
    iron_sword: int = 0  # Binary: 0 or 1

class Achievement(Enum):
    COLLECT_WOOD = 0
    PLACE_TABLE = 1
    ...
    MAKE_IRON_SWORD = 21
```

The reward function is calculated independently at each timestep
using these available factors:

- inventory_diff (Inventory): The change in the player's inventory
  between the current and previous timesteps.
- closest_blocks_changes (numpy.ndarray): The changes in distance to
  closest blocks of each type from the last timestep to the current
  timestep. Decreases in distance are positive.
- player_intrinsics_diff (jnp.ndarray): The changes in current
  intrinsic values (health, food, drink, energy) from the last
  timestep to the current timestep.

# Other Information
- This reward function is called independently at each timestep
- Each timestep's reward is calculated using only information from
  the current and previous timestep
- The completion criteria is a separate function
- No state can be stored between timesteps

# Skill
Given the following skill, design the reward function:
```
{skill_with_consumption}
```

# Design Approach
Return the raw factor value that measures the skill's gain.
Requirements and consumption are enforced elsewhere.

1. Identify the factor measuring the gain
2. Sparse reward: return the raw factor (can be positive, negative,
   or zero)
3. Dense reward: optional raw factor that provides denser feedback
   toward the gain, scaled by coefficient <= 0.01, or "NA".

```json
{
"sparse_reward_only_function": "return <factor>"
"dense_reward_function": "return <coefficient> * <dense_factor>"
}
```
\end{lstlisting}

\paragraph{Step 3b: Code Generation.}

\begin{lstlisting}[basicstyle=\ttfamily\scriptsize]
# [BlockType, Inventory, Achievement definitions as above]

# Here are example docstrings:

def task_is_done(inventory, inventory_diff, closest_blocks,
                 closest_blocks_prev, player_intrinsics,
                 player_intrinsics_diff, achievements, n):
    """
    Determines whether Task is complete by checking the primary gain
    from the gain dictionary.

    Args:
        inventory (Inventory): The player's current inventory
        inventory_diff (Inventory): The change in inventory
        closest_blocks (numpy.ndarray): Shape (len(BlockType), 2, K).
            Default values are (30, 30) for unseen entries.
        closest_blocks_prev (numpy.ndarray): Previous timestep blocks
        player_intrinsics (jnp.ndarray): Length 4 array (health,
            food, drink, energy)
        player_intrinsics_diff (jnp.ndarray): Change in intrinsics
        achievements (jnp.ndarray): Shape (22,) boolean array
        n (int): The count parameter for count-based gains

    Returns:
        bool: True if the primary gain condition is satisfied
    """
    return TODO


def task_reward(inventory_diff, closest_blocks, closest_blocks_prev,
                player_intrinsics_diff, achievements_diff,
                health_penalty):
    """
    Calculates the reward based on changes in inventory and other
    factors.

    Args:
        inventory_diff (Inventory): Change in inventory
        closest_blocks (numpy.ndarray): Current closest blocks
        closest_blocks_prev (numpy.ndarray): Previous closest blocks
        health_penalty (float): Penalty for losing health
        player_intrinsics_diff (jnp.ndarray): Change in intrinsics
        achievements_diff (jnp.ndarray): Achievements completed this
            timestep

    Returns:
        float: Reward for RL agent

    Note:
        The task reward should be two parts:
          1. Sparse reward
          2. Dense reward
        Make sure to disable (2) if (1) is triggered:
          sparse_reward + (sparse_reward == 0.0) * dense_reward
    """
    return TODO + health_penalty


def task_network_number():
    """
    Returns the network index corresponding to the skill.
    Returns:
        int: Network index
    """
    return TODO

Given the above documentation, implement the functions for the
subtask with the following details:
```json
{skill_with_consumption}
```

The sparse and dense rewards are:
```
Sparse: {sparse_reward}
Dense: {dense_reward}
```

## Implementation Guidelines:

For `task_is_done`:
- Identify the main gain entry from the "gain" dictionary
- Check if the current inventory amount >= n (the target amount)
- Return True when the target amount is reached, False otherwise
- If a skill is ephemeral, use closest_blocks or achievements

For `task_reward` and `task_network_number`:
- Follow the existing reward structure and network numbering

The code you write should be able to be jax compiled, no if
statements. Use jnp.where for conditionals.
No need to retype BlockType, Inventory, and Achievement.
Return all three functions in a single code block.
\end{lstlisting}

\newpage
\subsection{Adaptations for Full Craftax}
\label{sec:app-prompt-craftax}

The full Craftax environment extends Craftax-Classic with dungeon exploration across multiple floors. The prompt structure remains identical; the key adaptation is a two-step floor analysis process for prerequisite inference.

\paragraph{Floor Analysis Prompt.}
Before determining requirements, the LLM analyzes the floor context:

\begin{lstlisting}[basicstyle=\ttfamily\scriptsize]
You are analyzing the context for a skill in Craftax to determine
reasonable equipment expectations.

# Skill
```json
{skill_json}
```

# Tutorial Context
```
{tutorial_text}
```

# Task
1. **Identify Floor**: Look at `floor_availability`. If multiple
   floors are listed, consider the **minimum** (lowest number) floor
   where this skill can be performed.
   - Floor 0: Overworld
   - Floor 1: Dungeon
   - Floor 2: Gnomish Mines
   - Floor 3: Sewers
   - ...and so on.

2. **Determine Reasonable Equipment**:
   - Read the tutorial sections corresponding to the skill's floor
     and all previous floors.
   - **Resource Analysis**: What resources does the tutorial
     explicitly state are available, abundant, or rare on these
     floors? (e.g. look for phrases like "rich in ores" or "wood is
     rarer").
   - **Danger Analysis**: What enemies are present? Does the tutorial
     recommend specific weapons, armor tiers, or damage types (e.g.
     "requires fire damage")?
   - **Conclusion**: Based *strictly* on the tutorial evidence, what
     is the reasonable minimum tier of tools (Wood/Stone/Iron/Diamond)
     the player should have equipped to survive and function on this
     floor?

3. **Output**: Provide a concise summary of the floor context and the
   deduced tool tier for this skill.
\end{lstlisting}

\paragraph{Modified Prerequisite Prompt.}
The prerequisite inference prompt is then augmented with the floor analysis:

\begin{lstlisting}[basicstyle=\ttfamily\scriptsize]
# Floor & Equipment Analysis
```
{floor_analysis_output}
```

[... same structure as Phase 2 prompt ...]

## Step 2: Determine Requirements and Consumption
Based on the tutorial information AND the **Floor & Equipment
Analysis** above:

**Requirements**: Everything needed for safe and successful execution.
- The tier of tools should be informed by the Floor & Equipment
  Analysis above.
- Include location requirements using "level:player_level" if
  floor_availability is not empty.

[...]

## Step 3: Determine Ephemeral Status
[...]
Floor transitions are NOT ephemeral.
\end{lstlisting}

This two-step process ensures equipment requirements reflect the actual difficulty and resource availability of each dungeon floor rather than defaulting to conservative estimates.

\newpage
\subsection{Trajectory Analysis}
\label{sec:app-prompt-trajectory}

During training, SCALAR analyzes trajectories to refine skill specifications. This occurs in two scenarios: when training succeeds, to verify and correct requirements and consumption; and when training fails, to identify missing prerequisites.

\paragraph{Success Analysis.}
When a skill achieves sufficient success rate, the first successful trajectories are analyzed to verify the skill specification matches actual execution.

\begin{lstlisting}[basicstyle=\ttfamily\scriptsize]
You need to update a skill based on its execution trajectory.

Current Skill:
```
{skill_with_consumption}
```

Existing Skills (for requirements validation):
```
{skills_without_code}
```

Trajectory Data:
```
{example_trajectory}
```

## Task

Analyze the trajectory to determine what the skill actually required,
consumed, and gained, then express this in terms of `n` (the number
of times the skill is applied).

The trajectory shows a specific instance (e.g. n=1), but you need to
infer the general pattern.

**HOW TO READ THE TRAJECTORY:**
The trajectory contains:
1. **ACTION SUMMARY**: Shows the count of each action type performed
   during the entire trajectory
2. **TIMESTEPS WITH CHANGES**: Shows only the specific moments where
   inventory, achievements, or status changed

**YOUR ROLE IN TRAJECTORY ANALYSIS:**
Your job is to **correct factual errors** in requirements/consumption
based on what actually happened in the trajectory, NOT to judge
whether requirements are useful.

- **DO update** if the trajectory DISPROVES a requirement/consumption
  amount:
  - Example: Requirement says "wood: lambda n: 4*n" but trajectory
    shows crafting table only consumed 1 wood -> UPDATE to
    "lambda n: 1*n"
  - Example: Consumption says "stone: lambda n: 5*n" but trajectory
    shows only 3 stone was consumed -> UPDATE to "lambda n: 3*n"

- **DO NOT remove** requirements just because they weren't visibly
  used in this single trajectory:
  - Example: Requirement includes "wood_pickaxe" but trajectory
    doesn't explicitly show pickaxe usage -> KEEP the requirement
    (it may still be useful)
  - The initial skill definition already judged what's useful -
    trust that unless the trajectory proves it wrong

- **CRITICAL: PRESERVING TOOLS**:
  - Tools (Pickaxes, Swords) are DURABLE. They appear in
    `requirements` but NOT `consumption`.
  - **NEVER remove a tool requirement** (e.g. `wood_pickaxe`,
    `stone_sword`) unless the trajectory PROVES it was not needed
    (e.g. you punched the stone with your hand).
  - If the original skill required a Pickaxe, and the trajectory
    shows mining, **YOU MUST KEEP THE PICKAXE REQUIREMENT**.

- **DO update** gains to match what was actually achieved:
  - If trajectory shows achievement unlocked or items obtained,
    ensure gains reflect this

**IMPORTANT CONSTRAINTS:**
- Requirements are a SUPERSET of consumption: requirements include
  everything needed (both consumed and non-consumed resources),
  while consumption only includes what gets used up.
- Each value in requirements/consumption should be written as a
  Python lambda function string: "lambda n: a*n + b", where:
  - a = amount of resource consumed PER unit of gain (scales with n)
  - b = amount of resource required but NOT consumed (fixed amount)
  - Ask yourself: "Does this requirement scale with the number of
    times I apply the skill?"
    - If YES (scales with n): use "lambda n: a*n + 0" format
    - If NO (fixed amount): use "lambda n: 0*n + b" format
- Requirements do not support 'or'
- Each key in requirements/consumption must be a key in the gain of
  an existing skill.

# Formatting
```json
{
"skill_name": "",
"updated_requirements": {},
"updated_consumption": {},
"updated_gain": {}
}
```
\end{lstlisting}

\paragraph{Failure Analysis.}
When a skill fails to achieve sufficient success rate below 2\%, the LLM analyzes failure trajectories to identify missing prerequisites.

\begin{lstlisting}[basicstyle=\ttfamily\scriptsize]
Analyze why this skill failed (success rate < 2%) and identify
missing prerequisites.

Current Skill:
```
{skill_with_consumption}
```

Existing Skills:
```
{skills_without_code}
```

Knowledgebase (available skills):
```
{knowledge_base}
```

Failure Trajectory:
```
{failure_trajectory}
```

## Task

Identify what ADDITIONAL prerequisites would address the failure.
Keep existing requirements - only add what's missing.

**Requirements must be GAINS from knowledgebase skills.** Look at
the "gain" field of skills to find valid requirements.

**Intrinsic failures (health/food/drink/energy depletion):** Do NOT
add achievement prerequisites for survival behaviors (COLLECT_DRINK,
EAT_COW, WAKE_UP, etc.) - these only run once but survival requires
repeating them throughout the episode. Instead, add ITEM
prerequisites (wood_sword, stone, etc.) that help the agent survive
while learning these behaviors within the skill.

# Output (always return valid JSON)
```json
{
"skill_name": "",
"failure_analysis": "brief description of failure cause",
"suggested_prerequisites": [],
"updated_requirements": {},
"updated_consumption": {},
"updated_gain": {}
}
```
\end{lstlisting}

The failure analysis prompt emphasizes that new requirements must come from gains of existing skills in the knowledge base, ensuring the dependency graph remains well-formed. The distinction between intrinsic failures from survival depletion and capability failures from missing tools or resources guides the LLM toward adding actionable prerequisites rather than one-time achievements that cannot sustain the agent through extended skill execution.


\end{document}